%% file: main.tex
\documentclass{article} 
\usepackage{iclr2026_conference,times}

\input{math_commands.tex}

\usepackage[utf8]{inputenc} 
\usepackage[T1]{fontenc}    
\usepackage{hyperref}       
\usepackage{url}            
\usepackage{booktabs}       
\usepackage{amsfonts}       
\usepackage{nicefrac}       
\usepackage{microtype}      
\usepackage{xcolor}         

 \usepackage{graphicx}
\usepackage{algorithm}
\usepackage{algpseudocode} 
\usepackage{amsmath}
\usepackage[english]{babel}
\usepackage{subcaption}
\usepackage{multirow}
\usepackage{xspace}
\usepackage{bbm}
\usepackage{cleveref}
\crefformat{equation}{eq.~(#2#1#3)}
\usepackage{wrapfig}
\usepackage{enumitem}
\usepackage{adjustbox}

\newcommand{\mr}[2]{\multirow{#1}{*}{#2}}
\newcommand{\mc}[3]{\multicolumn{#1}{#2}{#3}}

\newcommand{\amazon}{\texttt{rel-amazon}\xspace}
\newcommand{\fone}{\texttt{rel-f1}\xspace}

\newcommand{\stack}{\texttt{rel-stack}\xspace}
\newcommand{\relbench}{\textsc{RelBench}\xspace}

\newcommand{\salt}{\texttt{rel-salt}\xspace}
\newcommand{\ratebeer}{\texttt{rel-ratebeer}\xspace}
\newcommand{\mimic}{\texttt{rel-mimic}\xspace}
\newcommand{\arxiv}{\texttt{rel-arxiv}\xspace}

\newcommand{\valtimestamp}{\textsc{val\_timestamp}\xspace}
\newcommand{\testtimestamp}{\textsc{test\_timestamp}\xspace}

\title{\relbench v2: A Large-Scale Benchmark and Repository for Relational Data}

\author{Justin Gu$^1$,
Rishabh Ranjan$^1$,
Charilaos Kanatsoulis$^1$,
Haiming Tang$^2$,
Martin Jurkovic$^3$,\\
\textbf{Valter Hudovernik$^4$,
Mark Znidar$^5$,
Pranshu Chaturvedi$^1$,
Parth Shroff$^1$,
Fengyu Li$^1$,}\\
\textbf{Jure Leskovec$^1$}\\
$^1$Stanford University,
$^2$National University of Singapore,
$^3$University of Ljubljana,
$^4$Kumo AI,\\
$^5$University of Oxford\\
\texttt{\{justingu,ranjanr,jure\}@stanford.edu} \\
Website: \textcolor{cyan}{\url{https://relbench.stanford.edu}}
}

\iclrfinalcopy 
\begin{document}

\maketitle

\begin{abstract}
Relational deep learning (RDL) has emerged as a powerful paradigm for learning directly on relational databases by modeling entities and their relationships across multiple interconnected tables. As this paradigm evolves toward larger models and relational foundation models, scalable and realistic benchmarks are essential for enabling systematic evaluation and progress. In this paper, we introduce \relbench v2, a major expansion of the \relbench benchmark for RDL. \relbench v2 adds four large-scale relational datasets spanning scholarly publications, enterprise resource planning, consumer platforms, and clinical records, contributing over 22 million rows across 29 tables and growing the benchmark to 11 datasets. We further introduce autocomplete tasks, a new class of predictive objectives that require models to infer missing attribute values directly within relational tables while respecting temporal constraints, expanding beyond traditional forecasting tasks constructed via SQL queries. In addition, \relbench v2 expands beyond its native datasets by integrating external benchmarks and evaluation frameworks: we translate event streams from the Temporal Graph Benchmark into relational schemas for unified relational–temporal evaluation, interface with ReDeLEx to provide uniform access to 70+ real-world databases suitable for pretraining, and incorporate 4DBInfer datasets and tasks to broaden multi-table evaluation coverage. Experimental results demonstrate that RDL models consistently outperform single-table baselines across autocomplete, forecasting, and recommendation tasks, highlighting the importance of modeling relational structure explicitly. 

\end{abstract}

\section{Introduction} \label{sec:intro}
Relational databases are the primary storage abstraction for structured data across enterprise, scientific, and healthcare systems. They organize information across multiple tables interconnected via primary–foreign key relationships, capturing rich structural and temporal dependencies between entities. Forecasting tasks such as predicting customer churn, estimating sales, recommending products, and anticipating system or patient outcomes are central to real-world decision making and have driven significant interest in applying machine learning to relational data.


Relational deep learning (RDL) \citep{relbench, rdl} has emerged as a powerful paradigm for learning directly on relational databases, reducing the human effort and engineering complexity of traditional machine learning pipelines that rely on manually flattening relational schemas into single tables through feature engineering and aggregation. Instead, RDL treats relational databases as heterogeneous graphs and applies graph neural networks \citep{relbench,chen2025relgnn} and other relational representation learning architectures \citep{dwivedi2025relational,dwivedi2025relational2} to model entities and their relationships directly. RDL models have demonstrated state-of-the-art performance in forecasting tasks over relational data.

At the same time, the machine learning landscape has shifted toward foundation models, which are pretrained on large and diverse datasets to learn transferable representations. This paradigm has transformed natural language processing and computer vision, and has recently extended to tabular data, where pretrained models achieve strong performance across predictive tasks \citep{tabpfn,tabpfnv2,tabicl}. However, these approaches focus on single-table data and do not capture the multi-table structure of relational databases. To address this limitation, recent efforts~\citep{kumorfm, griffin, ranjan2025relational} develop foundation models for relational databases, leveraging RDL architectures to model entities and their interdependencies across tables. Given their ubiquity and structural richness, relational databases represent a natural next frontier for foundation models, motivating benchmarks that capture realistic relational structure and predictive tasks.

To support machine learning research for forecasting on relational databases, \citet{relbench} introduced  \relbench, the first benchmark for RDL. \relbench v1 provided a collection of real-world relational databases along with forecasting tasks that require predicting future outcomes such as entity churn, sales, or recommendations. These tasks enabled standardized evaluation of relational learning models and demonstrated the effectiveness of RDL compared to traditional approaches.

In this work, we introduce \relbench v2, a significant expansion of the benchmark that presents new datasets, new task types, and a new paradigm for relational prediction. \relbench v2 adds four new large-scale relational datasets spanning diverse domains, increasing the total number of datasets to eleven. These include \arxiv, a scholarly publication database capturing papers, authors, categories, and citation relationships; \salt, an enterprise resource planning dataset modeling sales orders and business workflows; \ratebeer, a consumer platform dataset containing user interactions, product information, and reviews; and \mimic, a clinical dataset derived from electronic health records. Together, these datasets introduce new relational structures, domains, and predictive challenges, and collectively contain over 22 million rows across 29 tables. 

In addition to expanding the datasets, \relbench v2 introduces new predictive tasks and extends the benchmark with multiple strategic integrations of diverse external benchmarks and diagnostic frameworks. Most notably, \relbench v2 introduces autocomplete tasks, where the objective is to predict values of existing columns in relational tables at a given timestamp, requiring models to infer missing values from relational and temporal context while preventing information leakage. In addition, as subsequent efforts following \relbench v1 have expanded the scale and diversity of relational benchmarks, we integrate several of these into \relbench v2. These include temporal interaction datasets from the Temporal Graph Benchmark (TGB) \citep{rossi2020temporal}, widespread RDL evaluation on 70+ relational databases via ReDeLEx \citep{peleska2025redelex}, and the 4D design space for graph-centric relational modeling from 4DBInfer \citep{wang20244dbinfer}. Additional discussion on RDL and relational foundation models can be found in Appendix~\ref{sec:related-work}.

\section{Overview and Design} \label{sec:overview-design}

\relbench v1 \citep{relbench} laid the framework for creating a benchmark for deep learning
on relational databases. The benchmark consists of two key components: a collection of diverse real-world relational databases, and each database's corresponding set of realistic predictive tasks. 

\begin{itemize}[leftmargin=0.6cm,itemsep=3pt, parsep=0pt]
    \item \textbf{Relational databases}, consisting of a set of tables connected via primary–foreign key
    relationships. Tables store diverse information about entities, and some include time columns
    indicating when rows are created (e.g., transaction date). Each database is associated with fixed
    \valtimestamp and \testtimestamp cutoffs: models are trained on data up to \valtimestamp, validated
    on rows between \valtimestamp and \testtimestamp, and tested on rows after \testtimestamp. Data
    beyond the test cutoff is hidden during inference to prevent test-time leakage \citep{kapoor2023leakage},
    using the temporal neighbor sampling strategy of \citet{rdl}.

    \item \textbf{Predictive tasks} are defined per database via a training table \citep{rdl}. Each training
    table specifies an entity ID, a seed time, and target labels. The seed time determines when the
    prediction is made and filters out future information. Importantly, the \valtimestamp and
    \testtimestamp cutoffs are shared across all tasks within a dataset, enabling multi-task learning
    and pre-training across predictive tasks defined on the same relational database.
\end{itemize}

\textbf{Autocomplete tasks}: \relbench v2 introduces autocomplete tasks, a new paradigm of predictive tasks. These tasks allow for making predictions on the existing columns within tables in the dataset, as opposed to the previous \textbf{forecasting tasks} that predict on target labels constructed via SQL queries. However, like forecasting tasks, autocomplete tasks are still temporal in nature. Autocomplete tasks can be thought of as adding a new row to the database, filling some columns, and then trying to predict the remaining columns without access to future data. For each autocomplete task, we define a fixed set of observed columns and a target column to predict, where models receive the observed values for a row at a given seed time and must infer the target using only relational and temporal context available up to that time.

Autocomplete tasks expand the utility of the benchmark and widen the scope of real-world RDL applications; successfully predicting on autocomplete tasks requires models to deeply understand the relational context of the data. They have many real-world applications. In fact, these tasks were inspired by the SALT \citep{sap-salt} sales order autocomplete task (see Figure~\ref{fig:autocomplete-example}), where the SAP S/4HANA Sales Order user interface recommends a payment category based on answers to other data fields and contextual knowledge from the relational schema.

A key consideration for autocomplete tasks involves preventing information leakage, as some columns in a table may be highly correlated. Therefore, designing autocomplete tasks requires dropping the columns that correlate with the specified target column. To prevent information leakage, we manually identify and remove columns that are highly correlated with the target column based on domain knowledge and inspection of the training data. For instance, in the \texttt{review-rating} task for the \amazon dataset, we must drop the 'review\_text' column, which otherwise would provide intertwined signals with the review ratings. This prevents such interdependent columns from guiding predictions on the target column, hence preserving the authenticity of each predictive task and maintaining its real-world applicability.

\textbf{RDL implementation}: \relbench v2 utilizes the same RDL framework defined by \citet{relbench}. To reiterate, we first encode raw row-level data into initial node embeddings via PyTorch Frame \citep{hu2024pytorch}, specifically with the ResNet tabular model \citep{gorishniy2021revisiting}. We perform temporal-aware subgraph sampling \citep{rdl} around each entity node at a given seed time, where the embeddings are passed into a heterogeneous GraphSAGE model \citep{hamilton2017inductive, fey2019fast} with sum-based neighbor aggregation to iteratively update node embeddings. Finally, task-specific prediction heads turn output embeddings into predictions.

The rest of this paper is organized as follows. Section~\ref{sec:datasets} describes the new \relbench relational databases. Sections~\ref{sec:autocomplete-tasks} and \ref{sec:forecasting-tasks} introduce the new autocomplete and forecasting predictive tasks for each \relbench dataset, including results from benchmarking our RDL implementation against baselines. Finally, Section~\ref{sec0:tgb_relbench} describes the expansion of the \relbench ecosystem through the integration of external datasets and multi-dimensional RDL benchmarking tools.

\section{\relbench Datasets} \label{sec:datasets}

In \relbench v2, we introduce four new datasets, bringing the total number of datasets in the benchmark to eleven. These new datasets expand \relbench into new domains such as scholarly citations and enterprise operations, widening the breadth of data the benchmark covers and strengthening its position as a core benchmark for foundation models in RDL. Each dataset's predictive autocomplete and forecasting tasks are explained in more detail in Sections~\ref{sec:autocomplete-tasks} and \ref{sec:forecasting-tasks}, respectively. Detailed statistics for the new datasets can be found in Table~\ref{tab:stats_datasets}.

\begin{table}[t]
  \centering
  \caption{{\bf Statistics of new \relbench datasets.} Datasets vary significantly in the number of tables, total number of rows, and number of columns. In this table, we only count rows available for test inference, i.e., rows up to the test time cutoff. Note: \mimic shifts dates for patient privacy.
  }
  \renewcommand{\arraystretch}{1.1}
  \label{tab:stats_datasets}
  \scriptsize
  \vspace{-5pt}
  \resizebox{\textwidth}{!}{
    \input{tables/stats_datasets}
  }
  \vspace{-10pt}
\end{table}

\subsection{\arxiv}

The arXiv-physics dataset \citep{arxiv_physics_dataset} is a large-scale relational benchmark of over 222,000 research papers published between 2018 and 2023, designed to expand \relbench into the domain of scholarly network analysis. It captures the complex evolution of scientific research through 1.5 million directed citation links, paper-author relationships mapped via unique ORCID identifiers, and a hierarchical taxonomy of 53 physics categories. By modeling these dense many-to-many relations between 143,000 authors and their respective research areas, the dataset provides a rich structure for evaluating RDL models within the academic citation ecosystem.

\subsection{\salt}

The Sales Autocompletion Linked Business Tables (SALT) \citep{sap-salt} database, released by SAP AI Research, provides an authentic relational dataset of end-to-end business transactions from an enterprise resource planning (ERP) system. Centered on sales document headers and line items linked to customer and address master data, SALT models internal enterprise workflows including sales offices, shipping points, and payment terms. Predictive tasks focus on multiclass classification of operational variables in real-world supply chain and order fulfillment settings. By contributing minimally-processed industry data, SALT offers a unique business perspective to \relbench.

\subsection{\ratebeer}

The RateBeer dataset provides over two decades of user interactions across distinct tables for beers, places, users, and brewers. Linked through well-defined foreign keys, these attribute-rich tables contain over 30 columns of multi-modal features—including text, categorical, and temporal data—while interaction tables offer granular feedback through multi-aspect sub-scores and textual reviews. \ratebeer contributes a dataset with strong potential for capturing user preferences in multiple ways; by mapping users to beers, the dataset provides powerful signals for modeling preferences via both explicit rating scores and implicit "Favorites" lists.

\subsection{\mimic}

The Medical Information Mart for Intensive Care IV (MIMIC-IV) \citep{johnson2024mimic} is a large, deidentified electronic health record (EHR) dataset containing clinical data from patients at the Beth Israel Deaconess Medical Center. Designed to support clinical research and machine learning, the \relbench implementation allows for deep customization, including parameters to limit patient or table counts, drop specific columns, and filter by features like age. While the standard \relbench download utilizes a subset of 20,000 patients, the dataset also supports integration with Google BigQuery for accessing the full MIMIC-IV v3.1 data. Due to the sensitive nature of real-world medical data, users must obtain proper credentials through \href{https://physionet.org/content/mimiciv/3.1/}{PhysioNet} to access the dataset. MIMIC-IV also applies patient-level date shifting for privacy, offsetting all timestamps by a random amount so dates appear in the 2110–2200 range rather than the original 2008–2022 collection period.

\section{Autocomplete tasks} \label{sec:autocomplete-tasks}

\relbench v2 introduces 23 autocomplete tasks for both existing and new datasets, grouped into two task types: autocomplete classification (Section~/ref{sec:autocomplete-classification}) and autocomplete regression (Section~\ref{sec:autocomplete-regression}). Tasks are named based on the table and column used as the target labels for the predictions. A full list of autocomplete tasks is given in Table~\ref{tab:tasks_autocomplete}, with high-level descriptions given in Appendix~\ref{sec:task_info}.

\begin{table}[t]
  \centering
  \caption{{\bf Full list of new autocomplete tasks.} Autocomplete tasks aim to make predictions on existing columns in the dataset.}
  \label{tab:tasks_autocomplete}
  \renewcommand{\arraystretch}{1.1}
  \scriptsize
  \vspace{-5pt}
  \resizebox{\textwidth}{!}{
    \input{tables/tasks_autocomplete}
  }
  \vspace{-10pt}
\end{table}

\subsection{Autocomplete classification} \label{sec:autocomplete-classification}

Autocomplete classification tasks aim to predict labels for existing categorical columns in a dataset. Because categorical data often comes in both binary and multiclass situations, autocomplete tasks support both binary classification and multiclass classification. For binary classification, we use the ROC-AUC \citep{hanley1983method} metric for evaluation, where higher scores are better. For multiclass classification, we use accuracy as the evaluation metric, where higher is also better. As a baseline to compare our heterogeneous GraphSAGE model against, we utilize a LightGBM classifier baseline over the raw entity table features.

\textbf{Experimental results}. Classification results for the new autocomplete tasks are given in Table~\ref{tab:autocomplete_classification_main} for binary classification and Table~\ref{tab:autocomplete_multiclass_main} for multiclass classification. In both types of classification, RDL strongly outperforms the LightGBM baseline in all cases. On some tasks, such as \texttt{item-shippoint} from \salt or \texttt{badges-class} from \stack, RDL's high accuracy indicates that relational context gives highly informative signal to predict missing attributes.

For other tasks such as \texttt{sales-office} from \salt, comparison with the majority-class baseline suggests a strong class imbalance in the target labels, with the majority baseline already achieving very high accuracy. Despite this, the GNN matches or slightly improves upon the majority baseline while maintaining strong performance on other tasks, whereas LightGBM shows unstable behavior and substantially worse test performance, suggesting limited generalization. Overall, these results highlight the ability of RDL to exploit relational structure for autocomplete tasks, even in the presence of class imbalance or sparse feature information.

\begin{table}[t]
  \centering
  \caption{{\bf Autocomplete binary classification results on \relbench.} Binary classification uses the AUROC metric (higher is better). Best values are in bold. Standard baselines of random choice and majority class both correspond to AUROC values of approximately $50.00$, so we exclude them below. See Table~\ref{tab:autocomplete_classification_appendix} for standard deviations.}
  \label{tab:autocomplete_classification_main}
  \renewcommand{\arraystretch}{1.1}
    \vspace{-5pt}
    \begin{adjustbox}{max width=0.6\textwidth, center}
    \input{tables/results/autocomplete_classification_main}
    \end{adjustbox}
  \vspace{-10pt}
\end{table}

\begin{table}[t]
  \centering
  \caption{{\bf Autocomplete multiclass classification results on \relbench.} Multiclass classification uses the accuracy metric (higher is better). Best values are in bold. See Table~\ref{tab:autocomplete_multiclass_appendix} for std. devs.}
  \label{tab:autocomplete_multiclass_main}
  \renewcommand{\arraystretch}{1.1}
  \scriptsize
  \vspace{-5pt}
    \begin{adjustbox}{max width=0.65\textwidth, center}
    \input{tables/results/autocomplete_multiclass_main}
    \end{adjustbox}
\end{table}

\subsection{Autocomplete regression} \label{sec:autocomplete-regression}

Autocomplete regression tasks involve predicting numerical labels of an entity at a given seed time. For our evaluation metric, we use $R^2$, where higher values are better. We compare our RDL approach against four baselines. Global zero predicts zero for all entities. Global mean/median calculates the global mean/median label value for the target column in the training data and predicts that mean/median for every entity. Entity mean/median calculates the mean/median value with respect to each entity, and predicts that mean/median for the entity. LightGBM joins the entity table with the task table to get raw features from both, then learns a LightGBM \citep{ke2017lightgbm} regressor over those raw features to predict numerical targets.

\textbf{Experimental results}. Autocomplete regression results are reported in Table~\ref{tab:autocomplete_regression_main}. Additionally, MAE metrics for autocomplete regression tasks are reported in Table~\ref{tab:autocomplete_regression_appendix} in Appendix~\ref{sec:appendix_task_results}. Across most tasks, RDL achieves higher $R^2$ values, outperforming both feature-based and aggregation baselines, indicating that capturing relational context improves explanatory power.

\begin{table}[t]
  \centering
  \caption{{\bf Autocomplete regression results on \relbench.} Regression uses the $R^2$ metric (higher is better). Best values are in bold. See Table~\ref{tab:autocomplete_regression_r2_appendix} for standard deviations and Table~\ref{tab:autocomplete_regression_appendix} for MAE results.}
  \label{tab:autocomplete_regression_main}
  \renewcommand{\arraystretch}{1.1}
  \vspace{-5pt}
  \scriptsize
    \resizebox{\textwidth}{!}{
    \input{tables/results/autocomplete_regression_r2_main}
    }
  \vspace{-10pt}
\end{table}

\section{New forecasting tasks} \label{sec:forecasting-tasks}

Forecasting tasks were the original predictive tasks defined in \relbench v1, and version 2 introduces 13 new ones. Split into entity classification, entity regression, and recommendation (link prediction), these tasks use SQL queries to construct new task target columns. While classification and regression are entity-level tasks, recommendation tasks aim to predict the next temporal links between two sets of entities for a given link type, such as whether users will purchase a certain product.

We define forecasting tasks for the new datasets \arxiv, \ratebeer, and \mimic, and we add a new recommendation task for \fone. New forecasting tasks are shown in Table~\ref{tab:tasks_forecasting}.

\begin{table}[t]
  \centering
  \caption{\textbf{Full list of new forecasting tasks}. Forecasting tasks make predictions on new target columns created using SQL queries.}
  \label{tab:tasks_forecasting}
  \renewcommand{\arraystretch}{1.1}
  \scriptsize
  \vspace{-5pt}
  \resizebox{\textwidth}{!}{
    \input{tables/tasks_forecasting}
  }
  \vspace{-10pt}
\end{table}


\subsection{Entity Classification}

In \relbench v1, all entity-level classification tasks were binary classification. In \relbench v2, we broaden to the multiclass case with the first entity multiclass classification task, \arxiv's \texttt{author-category} task. For entity-level forecasting tasks, both binary and multiclass classification are evaluated with the same metrics as their autocomplete counterparts, with entity binary classification using ROC-AUC \citep{hanley1983method} and multiclass classification using accuracy (for both, higher is better). We again compare to a LightGBM classifier baseline over the raw entity table features, but here only information from the single entity table is used.

\textbf{Experimental results}. Entity classification results for the new forecasting tasks are given in Table~\ref{tab:entity_classification_main} for binary classification and Table~\ref{tab:entity_multiclass_main} for multiclass classification, with RDL outperforming the LightGBM baseline in all cases. Notably, RDL vastly outperforms LightGBM on the multiclass \texttt{author-category} task, where predicting an author’s research area benefits from aggregating relational signals from coauthorship, citation patterns, and publication context. This suggests that as classification tasks become more complex, leveraging relational context becomes more important.

\begin{table}[t]
  \centering
  \caption{{\bf Entity binary classification results on \relbench.} Binary classification uses the AUROC metric (higher is better). Best values are in bold. Standard baselines of random choice and majority class both correspond to AUROC values of approximately $50.00$, so we exclude them below. See Table~\ref{tab:entity_classification_appendix} for standard deviations.}
  \label{tab:entity_classification_main}
  \renewcommand{\arraystretch}{1.1}
  \vspace{-5pt}
  \scriptsize
    \begin{adjustbox}{max width=0.6\textwidth, center}
    \input{tables/results/entity_classification_main}
    \end{adjustbox}
\end{table}

\begin{table}[t]
  \centering
  \caption{{\bf Entity multiclass classification results on \relbench.} Multiclass classification uses the accuracy metric (higher is better). Best values are in bold. See Table~\ref{tab:entity_multiclass_appendix} for standard deviations.}
  \label{tab:entity_multiclass_main}
  \renewcommand{\arraystretch}{1.1}
  \scriptsize
  \vspace{-5pt}
    \begin{adjustbox}{max width=0.65\textwidth, center}
    \input{tables/results/entity_multiclass_main}
    \end{adjustbox}
  \vspace{-10pt}
\end{table}

\subsection{Entity Regression}

Entity-level regression tasks involve predicting numerical labels of an entity at a given seed time. Like for autocomplete regression, our evaluation metric is $R^2$, where higher values are better. We compare our RDL approach against essentially the same baselines as described in Section~\ref{sec:autocomplete-regression}. The only modification is that for the LightGBM baseline for entity regression, only the raw features from the single entity table are used to predict the numerical targets.

\textbf{Experimental results}. The entity regression results in Table~\ref{tab:entity_regression_main} show our RDL implementation outperforms the
baselines across all new forecasting regression tasks. RDL achieves higher $R^2$ values, indicating improved explanatory power when relational information is incorporated. Additionally, MAE metrics are reported in Appendix~\ref{sec:appendix_task_results} (Table~\ref{tab:entity_regression_appendix}), where RDL consistently achieves lower errors. These results suggest that relational modeling provides consistent benefits for numeric prediction at the entity level, even when the target variable is defined on a single entity table.

\begin{table}[t]
  \centering
  \caption{{\bf Entity regression results on \relbench.} Regression uses the $R^2$ metric (higher is better). Best values are in bold. See Table~\ref{tab:entity_regression_r2_appendix} for standard deviations and Table~\ref{tab:entity_regression_appendix} for MAE results.}
  \label{tab:entity_regression_main}
  \renewcommand{\arraystretch}{1.1}
  \scriptsize
  \vspace{-5pt}
    \begin{adjustbox}{max width=0.95\textwidth, center}
    \input{tables/results/entity_regression_r2_main}
    \end{adjustbox}
  \vspace{-10pt}
\end{table}

\subsection{Recommendation}

Recommendation tasks involve predicting, for each source entity and seed time, a ranked list of the top-$K$ target entities. This requires calculating pairwise scores between source and target entities.

We evaluate two GNN-based models. In GraphSAGE \citep{hamilton2017inductive}, source and target embeddings are learned via message passing, and pairwise scores are computed using their inner product, with training performed using the Bayesian Personalized Ranking loss \citep{rendle2012bpr}. In ID-GNN \citep{you2021identity}, target embeddings are passed through a source-specific MLP prediction head to produce pairwise scores, and the model is trained with cross-entropy loss \citep{you2021identity}.

We report Mean Average Precision (MAP) @$K$ (higher is better), with $K$ set per task. Baselines include Past Visit (ranks targets prior per-entity visit frequency), Global Popularity (ranks targets by overall frequency in the training data), and LightGBM \citep{ke2017lightgbm} (predicts source–target links using concatenated entity features, augmented with popularity and past-visit rank features).

\begin{table}[t]
  \centering
  \caption{{\bf Recommendation results on \relbench.} Recommendation uses the MAP metric, where higher values are better. Best values are in bold. See Table~\ref{tab:recommendation_appendix} for standard deviations.}
  \label{tab:recommendation_main}
  \renewcommand{\arraystretch}{1.1}
  \scriptsize
  \vspace{-5pt}
    \resizebox{\textwidth}{!}{
    \input{tables/results/recommendation_main}
    }
  \vspace{-10pt}
\end{table}

\textbf{Experimental results}. Results are given in Table~\ref{tab:recommendation_main}. In general, we observe that either the RDL implementation using GraphSAGE \citep{hamilton2017inductive}, or ID-GNN \citep{you2021identity} as the GNN component performs best, often by a very significant margin. ID-GNN excels in settings where predictions are highly entity-specific, whereas the plain GNN performs better when such specificity is less critical. This behavior reflects the inductive biases of the two models: GraphSAGE primarily captures structural and neighborhood-based patterns, while ID-GNN explicitly incorporates node identity information. Additionally, increasing the number of layers in the RDL models from two layers to four tended to yield improved performance across both GNN-based models, although the four-layer GraphSAGE model encountered CUDA memory errors on an 80GB Nvidia A100. These results highlight the multi-hop nature of recommendation tasks.

\section{Integrating External Benchmarks into \relbench} \label{sec0:tgb_relbench}

\relbench~\citep{relbench} introduced the first standardized benchmark for forecasting over relational databases, enabling end-to-end evaluation of RDL methods on real-world multi-table datasets. Subsequent efforts have expanded the scale and diversity of relational benchmarks. In addition to the new relational databases and tasks introduced in \relbench v2, we also extend \relbench with direct integration of external benchmarks and diagnostic frameworks. These include a suite of large-scale \emph{temporal interaction} datasets sourced from the Temporal Graph Benchmark (TGB) \citep{rossi2020temporal}, widespread evaluation of RDL models on 70+ relational databases via ReDeLex \citep{peleska2025redelex}, and a 4D benchmarking toolbox spanning multiple datasets, tasks, graph construction strategies, and predictive models from 4DBInfer \citep{wang20244dbinfer}.

\subsection{Temporal Graph Benchmark (TGB)} \label{sec:tgb_datasets}

The Temporal Graph Benchmark (TGB) is a benchmark centered on learning from time-stamped event streams (temporal edges), with evaluation protocols that enforce strict chronological generalization. By translating TGB datasets into the \relbench database and task abstraction, we enable direct comparisons between (i) \emph{temporal GNN} baselines that operate on event streams and (ii) \emph{relational deep learning} baselines that operate on a multi-table schema with explicit primary/foreign key structure. Following the principle of normalization in database theory, we translate each node and edge type into its own table. We focus on TGB datasets, excluding knowledge graphs that require additional adjustments, and naturally map the remaining datasets to relational event logs targeting the following downstream tasks:
(i) Dynamic Link Property Prediction (\texttt{tgbl-*}),
(ii) Dynamic Node Property Prediction (\texttt{tgbn-*}), and
(iii) Temporal Heterogeneous Graph Link Prediction (\texttt{thgl-*}).


The converted TGB datasets cover diverse domains and scales, from small bipartite interaction graphs to multi-relational, multi-entity temporal databases with tens of millions of events. Each dataset becomes a \relbench \texttt{Database} (parquet tables plus schema metadata) together with temporal cutoffs and tasks, enabling training and evaluation under the same leakage-safe conventions used elsewhere in \relbench. The dataset statistics can be found in Table~\ref{tab:stats_tgb_datasets}. Additional details about the \relbench TGB datasets and experiments can be found in App.~\ref{sec:tgb_relbench}. 


\begin{table}[t]                                                                                                                                                                                                                                         
    \centering                                                                                                                                                                                                                                               
    \caption{{\bf Statistics of TGB datasets translated into \relbench.}                                                                                                                                                                                     
    We report the relational size of each translated dataset as stored in parquet:                                                                                                                                                                           
    number of tables, total number of rows (summed across all tables, up to the test-time cutoff), and total number of columns (summed across all tables).                                                                                                   
    }                                                                                                                                                                                                                                                        
    \label{tab:stats_tgb_datasets}                                                                                                                                                                                                                           
    \begingroup                                                                                                                                                                                                                                              
    \scriptsize  
    \vspace{-5pt}
    \setlength{\tabcolsep}{3pt}                                                                                                                                                                                                                              
    \renewcommand{\arraystretch}{1.1}                                                                                                                                                                                                                        
    \begin{tabular}{llrrr}                                                                                                                                                                                                                                   
      \toprule                                                                                                                                                                                                                                               
      Task family & Dataset & \#Tables & \#Rows & \#Cols \\                                                                                                                                                                                                  
      \midrule                                                                                                                                                                                                                                               
      \mr{5}{\texttt{tgbl-*} (link)} &                                                                                                                                                                                                                       
        \texttt{tgbl-wiki-v2}   & 3  & 166,701    & 7  \\                                                                                                                                                                                                    
      & \texttt{tgbl-review-v2} & 2  & 5,226,177  & 6  \\                                                                                                                                                                                                    
      & \texttt{tgbl-coin}      & 2  & 23,447,972 & 6  \\                                                                                                                                                                                                    
      & \texttt{tgbl-comment}   & 2  & 45,309,297 & 6  \\                                                                                                                                                                                                    
      & \texttt{tgbl-flight}    & 2  & 67,187,713 & 6  \\                                                                                                                                                                                                    
      \midrule                                                                                                                                                                                                                                               
      \mr{4}{\texttt{tgbn-*} (node)} &                                                                                                                                                                                                                       
        \texttt{tgbn-trade}  & 5 & 934,072     & 14 \\                                                                                                                                                                                                       
      & \texttt{tgbn-genre}  & 5 & 20,858,841  & 14 \\                                                                                                                                                                                                       
      & \texttt{tgbn-reddit} & 5 & 43,669,153  & 14 \\                                                                                                                                                                                                       
      & \texttt{tgbn-token}  & 5 & 81,663,534  & 14 \\                                                                                                                                                                                                       
      \midrule                                                                                                                                                                                                                                               
      \mr{4}{\texttt{thgl-*} (hetero link)} &                                                                                                                                                                                                                
        \texttt{thgl-software} & 18 & 2,171,733   & 74 \\                                                                                                                                                                                                    
      & \texttt{thgl-forum}    & 4  & 23,910,523  & 12 \\                                                                                                                                                                                                    
      & \texttt{thgl-github}   & 18 & 23,356,342  & 74 \\                                                                                                                                                                                                    
      & \texttt{thgl-myket}    & 4  & 55,163,623  & 12 \\                                                                                                                                                                                                    
      \bottomrule                                                                                                                                                                                                                                            
    \end{tabular}                                                                                                                                                                                                                                       
    \endgroup                                                                                                                                                                                                                            
    \vspace{-10pt}                                                                                                                                                                                                                                           
\end{table}                                                                                                                                                                  

\subsection{Relational Deep Learning Exploration (ReDeLex)} \label{sec:redelex}

Relational Deep Learning Exploration (ReDeLEx)~\citep{peleska2025redelex} is a large-scale experimental framework for systematically evaluating Relational Deep Learning (RDL) on real-world relational databases. From the CTU Relational Learning Repository~\citep{motl2025ctupraguerelationallearning}, ReDeLEx integrates over 70 datasets into a unified pipeline that connects directly to SQL databases, infers attribute semantics, and represents relational schemas as heterogeneous graphs. These datasets span domains such as healthcare, government, education, sports, and business applications, vastly increasing \relbench's coverage of data. RDL models in ReDeLEx follow a modular design that combines attribute encoders, optional tabular models, graph neural network layers, and task-specific prediction heads. These models enable controlled comparisons across architectures such as linear GraphSAGE, Tabular ResNet–augmented GNNs, and Transformer-based models.

\subsection{4DBInfer} \label{sec:4dbinfer}

4DBInfer~\citep{wang20244dbinfer} is a large-scale benchmarking effort focused on predictive modeling over multi-table relational databases. 4DBInfer introduces an explicit four-dimensional evaluation framework: datasets, tasks, relational-to-graph construction strategies, and predictive model families. These are designed to expose how modeling choices across the full pipeline impact performance. While 4DBInfer explores this design space through extensive empirical comparisons, its datasets and task formulations provide a valuable foundation for unified evaluation. As part of \relbench~v2, we incorporate 7 4DBInfer datasets and 12 tasks, allowing \relbench to inherit the scale and diversity of 4DBInfer while enforcing consistent experimental conventions across relational, temporal, and graph-based learning settings.

\section{Conclusion}
In this work, we introduced \relbench v2, a major expansion of the \relbench benchmark for relational deep learning (RDL). \relbench v2 adds four large-scale relational datasets spanning academic, enterprise, consumer, and clinical domains, substantially increasing the scale and diversity of real-world relational data. We further introduced autocomplete tasks, a new class of predictive objectives that require models to infer missing attribute values directly within relational tables under temporal constraints, complementing traditional forecasting and recommendation tasks. In addition, we integrated temporal interaction datasets from the Temporal Graph Benchmark (TGB), and relational databases from Relational Deep Learning Exploration (ReDeLEx) and 4DBInfer, enabling unified evaluation across relational and temporal learning settings. Experimental results show that RDL models consistently outperform single-table baselines across autocomplete, forecasting, and recommendation tasks, highlighting the importance of modeling relational structure explicitly. \relbench v2 provides a scalable and realistic benchmark to support the development and evaluation of RDL systems and relational foundation models.

\bibliography{iclr2026_conference}
\bibliographystyle{iclr2026_conference}

\newpage
\appendix
\section{Related Work} \label{sec:related-work}

\textbf{Relational deep learning (RDL).} RDL studies how to train neural models directly on relational databases by leveraging their multi-table structure. RDL represents a relational database as a heterogeneous graph, where rows correspond to entities and foreign-key relationships define edges between them \citep{rdl}. Early works applied graph neural networks to such relational graphs and demonstrated substantial improvements over feature-engineered baselines on forecasting, recommendation, and prediction tasks \citep{relbench, chen2025relgnn}. More recent approaches have explored transformer-based architectures to better capture long-range and higher-order dependencies across tables \citep{dbformer, dwivedi2025relational}, as well as positional encoding methods designed to improve representation learning on relational graphs \citep{kanatsoulis2025learningefficientpositionalencodings}.


\textbf{Foundation models for tabular and relational data.} Recent tabular foundation models demonstrate strong performance, including in-context learning~\citep{tabpfn, tabicl} and efficient fine-tuning~\citep{carte}. These models leverage supervised~\citep{tabpfn, tabpfnv2} or self-supervised~\citep{portal, carte} pretraining on real and synthetic tabular datasets. Extending such models to relational databases is challenging due to the presence of multiple tables connected via foreign-key relationships. To address this, relational foundation models have recently been proposed. For example, \citet{kumorfm} introduce KumoRFM, a graph-transformer-based architecture capable of in-context learning and fine-tuning. Similarly, \citet{griffin} pretrain the Griffin model on both tabular and relational datasets, combining table-level encoders with graph neural networks for cross-table reasoning. More recent approaches operate directly at the cell level and use attention mechanisms to explicitly represent foreign-key relationships, enabling unified reasoning across the entire relational database without requiring intermediate aggregation \citep{ranjan2025relational}.
To circumvent real-data limitations for large-scale pretraining, recent works have explored generating privacy-preserving versions of real databases with diffusion models~\citep{hudovernik2025reldiff,ketata2025joint} as well as generating synthetic databases from scratch using random graphs and Structural Causal Models (SCMs)~\citep{kothapalli2026plurel}. In contrast, in \relbench v2 we collect a large number of realistic databases in a uniformly accessible manner.

\section{Dataset schemas} \label{sec:dataset_schemas}

\input{sections/appendix-dataset-schemas}

\newpage

\section{Additional Task Information} \label{sec:task_info}

\input{sections/appendix-task-descriptions}

\newpage

\section{Additional results and experiment details} \label{sec:appendix_task_results}

\input{sections/appendix-task-std}

\end{document}

%% file: math_commands.tex
\usepackage{amsmath,amsfonts,bm}

\def\eqref#1{equation~\ref{#1}}

\def\1{\bm{1}}

\DeclareMathAlphabet{\mathsfit}{\encodingdefault}{\sfdefault}{m}{sl}
\SetMathAlphabet{\mathsfit}{bold}{\encodingdefault}{\sfdefault}{bx}{n}



%% file: tables/stats_datasets.tex
\begin{tabular}{lllrrrrrr}
    \toprule
      \mr{2}{Name} & \mr{2}{Domain} &  \mr{2}{\#Tasks} & \mc{3}{c}{Tables} &  \mc{3}{c}{Timestamp (year-mon-day)} \\
       \cmidrule(lr){4-6}\cmidrule(lr){7-9}
      & &  & \#Tables & \#Rows & \#Cols & Start & Val & Test  \\
    \midrule
    \salt & Enterprise & 8 & 4 & 4,257,145 & 31 & 2018-01-02 & 2020-02-01 & 2020-07-01 \\
    \arxiv & Academic & 4 & 6 & 2,146,112 & 21 & 2018-01-01 & 2022-01-01 & 2023-01-01 \\
    \ratebeer & Consumer & 8 & 13 & 13,787,005 & 221 & 2000-04-02 & 2018-09-01 & 2020-01-01 \\
    \mimic & Medical & 1 & 6 & 2,424,751 & 54 & 2110-01-14 & 2168-09-06 & 2180-10-26 \\

   \multicolumn{2}{c}{Total} & 21 & 29 &  22,615,013 & 327 & / & / & / \\
    \bottomrule
\end{tabular}

%% file: tables/tasks_autocomplete.tex
\begin{tabular}{lllrrrrrr}
    \toprule
      \mr{2}{Dataset} &
      \mr{2}{Task name} &
      \mr{2}{Task type} &
      \mc{3}{c}{\#Rows of training table} &
      \mr{2.5}{\shortstack{\#Unique\\Entities}} &
      \mr{2.8}{\shortstack{\%train/test\\Entity\\Overlap}} &
      \mr{2.5}{\shortstack{\#Dst\\Entities}} \\
      \cmidrule(lr){4-6}
      & & & Train & Validation & Test & & & \\
    \midrule

    \mr{1}{rel-amazon}
    & review-rating & auto-reg  & 11,822,796 & 806,355 & 8,217,532 & 17,255,399 & 40.5 & -- \\
    \midrule

    \mr{2}{rel-avito}
    & searchstream-click & auto-bcls & 2,212,750 & 1,177,380 & 924,990 & 3,976,413 & 23.8 & -- \\
    & searchinfo-isuserloggedon & auto-bcls & 1,291,566 & 695,590 & 592,133 & 2,579,289 & 0.0  & -- \\
    \midrule

    \mr{3}{rel-event}
    & event\_interest-interested & auto-bcls & 14,442 & 536  & 420  & 14,992 & 94.5 & -- \\
    & event\_interest-not\_interested & auto-bcls & 14,442 & 536 & 420 & 14,992 & 94.5 & -- \\
    & users-birthyear & auto-reg  & 33,937 & 1,731 & 1,002 & 36,670 & 0.0  & -- \\
    \midrule

    \mr{2}{rel-f1}
    & results-position & auto-reg & 8,997 & 1,400 & 4,798 & 15,195 & 0.0 & -- \\
    & qualifying-position & auto-reg & 2,228 & 1,854 & 5,733 & 9,815 & 0.0 & -- \\
    \midrule

    \mr{1}{rel-hm}
    & transactions-price & auto-reg & 14,844,291 & 235,662 & 266,364 & 15,346,317 & 0.0 & -- \\
    \midrule

    \mr{1}{rel-ratebeer}
    & beer\_ratings-total\_score & auto-reg & 10,620,177 & 1,227,702 & 2,495,360 & 14,343,239 & 0.0 & -- \\
    \midrule

    \mr{8}{rel-salt}
    & item-plant & auto-mcls & 1,622,787 & 293,823 & 400,206 & 2,316,816 & 0.0 & -- \\
    & item-shippoint & auto-mcls & 1,622,787 & 293,780 & 398,536 & 2,315,103 & 0.0 & -- \\
    & item-incoterms & auto-mcls & 1,622,787 & 293,891 & 402,835 & 2,319,513 & 0.0 & -- \\
    & sales-office & auto-mcls & 340,491 & 71,474 & 88,942 & 500,907 & 0.0 & -- \\
    & sales-group & auto-mcls & 340,491 & 70,224 & 83,193 & 493,908 & 0.0 & -- \\
    & sales-payterms & auto-mcls & 340,491 & 71,472 & 88,831 & 500,794 & 0.0 & -- \\
    & sales-shipcond & auto-mcls & 340,491 & 71,398 & 88,422 & 500,311 & 0.0 & -- \\
    & sales-incoterms & auto-mcls & 340,491 & 71,470 & 88,925 & 500,886 & 0.0 & -- \\
    \midrule

    \mr{1}{rel-stack}
    & badges-class & auto-mcls & 448,358 & 15,105 & 127,370 & 590,833 & 0.0 & -- \\
    \midrule

    \mr{4}{rel-trial}
    & studies-enrollment & auto-reg  & 233,072 & 14,470 & 23,430 & 270,972 & 0.0 & -- \\
    & studies-has\_dmc & auto-bcls & 202,840 & 11,983 & 18,944 & 233,767 & 0.0 & -- \\
    & eligibilities-adult & auto-bcls & 234,366 & 14,470 & 23,430 & 272,266 & 0.0 & -- \\
    & eligibilities-child & auto-bcls & 234,366 & 14,470 & 23,430 & 272,266 & 0.0 & -- \\
    \bottomrule
\end{tabular}

%% file: tables/results/autocomplete_classification_main.tex
\begin{tabular}{lllrr}
\toprule
Dataset & Task & Split & LightGBM & GNN \\
\midrule
\multirow[c]{4}{*}{\texttt{rel-avito}} & \multirow[c]{2}{*}{\texttt{searchinfo-isuserloggedon}} & Val & $59.09$ & $\bm{82.57}$ \\
 &  & Test & $50.00$ & $\bm{73.00}$ \\
\cmidrule{2-5}
 & \multirow[c]{2}{*}{\texttt{searchstream-click}} & Val & $\bm{68.33}$ & $50.39$ \\
 &  & Test & $49.92$ & $\bm{55.92}$ \\
\cmidrule{1-5} \cmidrule{2-5}
\multirow[c]{4}{*}{\texttt{rel-event}} & \multirow[c]{2}{*}{\texttt{event\_interest-interested}} & Val & $51.25$ & $\bm{54.16}$ \\
 &  & Test & $49.57$ & $47.64$ \\
\cmidrule{2-5}
 & \multirow[c]{2}{*}{\texttt{event\_interest-not\_interested}} & Val & $51.98$ & $49.74$ \\
 &  & Test & $52.88$ & $\bm{60.40}$ \\
\cmidrule{1-5} \cmidrule{2-5}
\multirow[c]{6}{*}{\texttt{rel-trial}} & \multirow[c]{2}{*}{\texttt{eligibilities-adult}} & Val & $58.10$ & $\bm{94.91}$ \\
 &  & Test & $50.00$ & $\bm{93.73}$ \\
\cmidrule{2-5}
 & \multirow[c]{2}{*}{\texttt{eligibilities-child}} & Val & $59.78$ & $\bm{85.91}$ \\
 &  & Test & $50.00$ & $\bm{87.25}$ \\
\cmidrule{2-5}
 & \multirow[c]{2}{*}{\texttt{studies-has\_dmc}} & Val & $76.47$ & $\bm{78.21}$ \\
 &  & Test & $50.00$ & $\bm{75.72}$ \\
\cmidrule{1-5} \cmidrule{2-5}
\multicolumn{2}{c}{\multirow{2}{*}{Average}} & Val & $60.71$ & $\bm{70.84}$ \\
 &  & Test & $50.34$ & $\bm{70.52}$ \\
\bottomrule
\end{tabular}

%% file: tables/results/autocomplete_multiclass_main.tex
\begin{tabular}{lllrrrr}
\toprule
Dataset & Task & Split & Random & Majority & LightGBM & GNN \\
\midrule
\multirow[c]{17}{*}{\texttt{rel-salt}} & \multirow[c]{2}{*}{\texttt{item-incoterms}} & Val & $34.49$ & $66.46$ & $66.43$ & $\bm{80.23}$ \\
 &  & Test & $30.33$ & $58.05$ & $58.05$ & $\bm{69.36}$ \\
\cmidrule{2-7}
 & \multirow[c]{2}{*}{\texttt{item-plant}} & Val & $33.19$ & $60.95$ & $60.97$ & $\bm{99.70}$ \\
 &  & Test & $32.38$ & $59.69$ & $59.69$ & $\bm{99.46}$ \\
\cmidrule{2-7}
 & \multirow[c]{2}{*}{\texttt{item-shippoint}} & Val & $8.20$ & $2.34$ & $4.72$ & $\bm{98.54}$ \\
 &  & Test & $6.53$ & $1.99$ & $5.67$ & $\bm{98.39}$ \\
\cmidrule{2-7}
 & \multirow[c]{2}{*}{\texttt{sales-group}} & Val & $0.90$ & $0.86$ & $0.70$ & $\bm{18.43}$ \\
 &  & Test & $0.85$ & $0.75$ & $0.94$ & $\bm{15.76}$ \\
\cmidrule{2-7}
 & \multirow[c]{2}{*}{\texttt{sales-incoterms}} & Val & $31.83$ & $61.00$ & $60.53$ & $\bm{69.07}$ \\
 &  & Test & $29.39$ & $56.63$ & $56.63$ & $\bm{62.23}$ \\
\cmidrule{2-7}
 & \multirow[c]{2}{*}{\texttt{sales-office}} & Val & $50.01$ & $\bm{99.91}$ & $99.90$ & $99.91$ \\
 &  & Test & $49.71$ & $\bm{99.88}$ & $59.93$ & $99.88$ \\
\cmidrule{2-7}
 & \multirow[c]{2}{*}{\texttt{sales-payterms}} & Val & $0.32$ & $0.65$ & $1.85$ & $\bm{39.88}$ \\
 &  & Test & $0.24$ & $0.47$ & $5.64$ & $\bm{37.47}$ \\
\cmidrule{2-7}
 & \multirow[c]{2}{*}{\texttt{sales-shipcond}} & Val & $16.49$ & $27.61$ & $31.92$ & $\bm{59.21}$ \\
 &  & Test & $15.64$ & $26.30$ & $4.91$ & $\bm{56.85}$ \\
\cmidrule{1-7} \cmidrule{2-7}
\multirow[c]{2}{*}{\texttt{rel-stack}} & \multirow[c]{2}{*}{\texttt{badges-class}} & Val & $11.59$ & $20.68$ & $1.93$ & $\bm{79.97}$ \\
 &  & Test & $10.49$ & $18.34$ & $2.51$ & $\bm{82.83}$ \\
\cmidrule{1-7} \cmidrule{2-7}
\multicolumn{2}{c}{\multirow{2}{*}{Average}} & Val & $20.78$ & $37.83$ & $36.55$ & $\bm{71.66}$ \\
 &  & Test & $19.51$ & $35.79$ & $28.22$ & $\bm{69.14}$ \\
\bottomrule
\end{tabular}

%% file: tables/results/autocomplete_regression_r2_main.tex
\begin{tabular}{lllrrrrrrr}
\toprule
Dataset & Task & Split & Zero & Mean & Median & Ent. Mean & Ent. Med. & LightGBM & GNN \\
\midrule
\multirow[c]{2}{*}{\texttt{rel-amazon}} & \multirow[c]{2}{*}{\texttt{review-rating}} & Val & $-20.848$ & $\bm{-0.006}$ & $-0.364$ & $-20.848$ & $-20.848$ & $-0.364$ & $-0.356$ \\
 &  & Test & $-22.579$ & $\bm{-0.014}$ & $-0.341$ & $-13.313$ & $-13.313$ & $-0.341$ & $-0.331$ \\
\cmidrule{1-10} \cmidrule{2-10}
\multirow[c]{2}{*}{\texttt{rel-event}} & \multirow[c]{2}{*}{\texttt{users-birthyear}} & Val & $-55012.323$ & $-0.047$ & $-0.216$ & $-55012.323$ & $-55012.323$ & $0.004$ & $\bm{0.008}$ \\
 &  & Test & $-64803.758$ & $-0.121$ & $-0.395$ & $-64803.758$ & $-64803.758$ & $-0.192$ & $\bm{-0.030}$ \\
\cmidrule{1-10} \cmidrule{2-10}
\multirow[c]{4}{*}{\texttt{rel-f1}} & \multirow[c]{2}{*}{\texttt{qualifying-position}} & Val & $-3.267$ & $-0.018$ & $-0.030$ & $-3.267$ & $-3.267$ & $\bm{0.153}$ & $0.015$ \\
 &  & Test & $-3.214$ & $-0.002$ & $-0.001$ & $-3.214$ & $-3.214$ & $-0.953$ & $\bm{0.015}$ \\
\cmidrule{2-10}
 & \multirow[c]{2}{*}{\texttt{results-position}} & Val & $-3.123$ & $-0.100$ & $-0.107$ & $-3.123$ & $-3.123$ & $0.283$ & $\bm{0.440}$ \\
 &  & Test & $-3.148$ & $-0.176$ & $-0.219$ & $-3.148$ & $-3.148$ & $-2.437$ & $\bm{0.394}$ \\
\cmidrule{1-10} \cmidrule{2-10}
\multirow[c]{2}{*}{\texttt{rel-hm}} & \multirow[c]{2}{*}{\texttt{transactions-price}} & Val & $-2.215$ & $-0.065$ & $-0.140$ & $-2.215$ & $-2.215$ & $-0.140$ & $\bm{0.725}$ \\
 &  & Test & $-2.329$ & $-0.075$ & $-0.159$ & $-2.329$ & $-2.329$ & $-0.160$ & $\bm{0.736}$ \\
\cmidrule{1-10} \cmidrule{2-10}
\multirow[c]{2}{*}{\texttt{rel-ratebeer}} & \multirow[c]{2}{*}{\texttt{beer\_ratings-total\_score}} & Val & $-23.411$ & $-0.015$ & $-0.004$ & $-23.411$ & $-23.411$ & $-0.004$ & $\bm{0.448}$ \\
 &  & Test & $-34.352$ & $-0.031$ & $-0.003$ & $-34.352$ & $-34.352$ & $-0.014$ & $\bm{0.394}$ \\
\cmidrule{1-10} \cmidrule{2-10}
\multirow[c]{2}{*}{\texttt{rel-trial}} & \multirow[c]{2}{*}{\texttt{studies-enrollment}} & Val & $-0.001$ & $\bm{-0.000}$ & $-0.001$ & $-0.001$ & $-0.001$ & $-0.001$ & $-0.000$ \\
 &  & Test & $-0.000$ & $\bm{-0.000}$ & $-0.000$ & $-0.000$ & $-0.000$ & $-0.000$ & $-0.000$ \\
\bottomrule
\end{tabular}

%% file: tables/tasks_forecasting.tex
\begin{tabular}{lllrrrrrr}
    \toprule
      \mr{2}{Dataset} &
      \mr{2}{Task name} &
      \mr{2}{Task type} &
      \mc{3}{c}{\#Rows of training table} &
      \mr{2.5}{\shortstack{\#Unique\\Entities}} &
      \mr{2.8}{\shortstack{\%train/test\\Entity\\Overlap}} &
      \mr{2.5}{\shortstack{\#Dst\\Entities}} \\
      \cmidrule(lr){4-6}
      & & & Train & Validation & Test & & & \\
    \midrule


    \mr{4}{rel-arxiv}
    & paper-citation & entity-bcls & 534,233 & 155,845 & 193,696 & 136,183 & 70.31 & -- \\
    & author-category & entity-mcls & 210,769 & 39,015 & 39,655 & 126,219 & 62.15 & -- \\
    & author-publication & entity-reg & 210,769 & 39,015 & 39,655 & 101,886 & 62.15 & -- \\
    & paper-paper-cocitation & recommendation & 246,341 & 71,257 & 82,033 & 94,289 & 60.57 & 138,688 \\
    \midrule



    \mr{1}{rel-f1}
    & driver-circuit-compete & recommendation & 2,649 & 27 & 27 & 786 & 40.74 & 19,044 \\
    \midrule


    \mr{1}{rel-mimic}
    & patient-iculengthofstay & entity-bcls & 13,816 & 2,699 & 2,445 & 13,816 & 0.00 & -- \\
    \midrule

    \mr{8}{rel-ratebeer}
    & beer-churn & entity-bcls & 2,470,686 & 92,367 & 79,927 & 516,368 & 45.46 & -- \\
    & user-churn & entity-bcls & 373,709 & 19,908 & 9,392 & 154,071 & 35.21 & -- \\
    & brewer-dormant & entity-bcls & 98,697 & 15,840 & 16,366 & 28,333 & 64.07 & -- \\
    & user-count & entity-reg & 373,709 & 19,908 & 9,392 & 154,071 & 35.21 & -- \\
    & user-beer-favorite & recommendation & 1,099 & 1,043 & 499 & 2,296 & 10.82 & 7,745 \\
    & user-beer-liked & recommendation & 150,322 & 5,681 & 2,783 & 35,010 & 58.53 & 170,964 \\
    & user-place-liked & recommendation & 38,444 & 547 & 351 & 7,425 & 81.77 & 46,814 \\
    \midrule

\end{tabular}

%% file: tables/results/entity_classification_main.tex
\begin{tabular}{lllrr}
\toprule
Dataset & Task & Split & LightGBM & GNN \\
\midrule
\multirow[c]{2}{*}{\texttt{rel-arxiv}} & \multirow[c]{2}{*}{\texttt{paper-citation}} & Val & $71.94$ & $\bm{82.45}$ \\
 &  & Test & $71.21$ & $\bm{82.50}$ \\
\cmidrule{1-5} \cmidrule{2-5}
\multirow[c]{2}{*}{\texttt{rel-mimic}} & \multirow[c]{2}{*}{\texttt{patient-iculengthofstay}} & Val & $53.64$ & $\bm{56.52}$ \\
 &  & Test & $51.81$ & $\bm{55.01}$ \\
\cmidrule{1-5} \cmidrule{2-5}
\multirow[c]{6}{*}{\texttt{rel-ratebeer}} & \multirow[c]{2}{*}{\texttt{beer-churn}} & Val & $81.90$ & $\bm{90.47}$ \\
 &  & Test & $76.21$ & $\bm{78.67}$ \\
\cmidrule{2-5}
 & \multirow[c]{2}{*}{\texttt{brewer-dormant}} & Val & $76.39$ & $\bm{82.10}$ \\
 &  & Test & $75.79$ & $\bm{80.51}$ \\
\cmidrule{2-5}
 & \multirow[c]{2}{*}{\texttt{user-churn}} & Val & $87.02$ & $\bm{96.85}$ \\
 &  & Test & $83.92$ & $\bm{94.27}$ \\
\cmidrule{1-5} \cmidrule{2-5}
\multicolumn{2}{c}{\multirow{2}{*}{Average}} & Val & $74.18$ & $\bm{81.68}$ \\
 &  & Test & $71.79$ & $\bm{78.19}$ \\
\bottomrule
\end{tabular}

%% file: tables/results/entity_multiclass_main.tex
\begin{tabular}{lllrrrr}
\toprule
Dataset & Task & Split & Random & Majority & LightGBM & GNN \\
\midrule
\multirow[c]{2}{*}{\texttt{rel-arxiv}} & \multirow[c]{2}{*}{\texttt{author-category}} & Val & $1.75$ & $8.83$ & $1.95$ & $\bm{52.63}$ \\
 &  & Test & $1.77$ & $9.09$ & $2.01$ & $\bm{50.74}$ \\
\bottomrule
\end{tabular}

%% file: tables/results/entity_regression_r2_main.tex
\begin{tabular}{lllrrrrrrr}
\toprule
Dataset & Task & Split & Zero & Mean & Median & Ent. Mean & Ent. Med. & LightGBM & GNN \\
\midrule
\multirow[c]{4}{*}{\texttt{rel-amazon}} & \multirow[c]{2}{*}{\texttt{item-ltv}} & Val & $-0.025$ & $-0.000$ & $-0.013$ & $-0.247$ & $-0.107$ & $0.002$ & $\bm{0.066}$ \\
 &  & Test & $-0.013$ & $-0.000$ & $-0.007$ & $0.030$ & $\bm{0.099}$ & $0.001$ & $0.032$ \\
\cmidrule{2-10}
 & \multirow[c]{2}{*}{\texttt{user-ltv}} & Val & $-0.084$ & $-0.003$ & $-0.084$ & $0.053$ & $0.095$ & $-0.084$ & $\bm{0.195}$ \\
 &  & Test & $-0.092$ & $-0.000$ & $-0.092$ & $0.143$ & $0.168$ & $-0.092$ & $\bm{0.172}$ \\
\cmidrule{1-10} \cmidrule{2-10}
\multirow[c]{2}{*}{\texttt{rel-arxiv}} & \multirow[c]{2}{*}{\texttt{author-publication}} & Val & $-1.579$ & $-0.012$ & $-0.259$ & $0.254$ & $0.236$ & $-0.259$ & $\bm{0.437}$ \\
 &  & Test & $-1.572$ & $-0.000$ & $-0.210$ & $-0.010$ & $-0.064$ & $-0.210$ & $\bm{0.249}$ \\
\cmidrule{1-10} \cmidrule{2-10}
\multirow[c]{2}{*}{\texttt{rel-avito}} & \multirow[c]{2}{*}{\texttt{ad-ctr}} & Val & $-0.238$ & $-0.002$ & $-0.095$ & $-0.224$ & $-0.224$ & $-0.032$ & $\bm{0.030}$ \\
 &  & Test & $-0.226$ & $-0.004$ & $-0.098$ & $-0.148$ & $-0.148$ & $-0.039$ & $\bm{-0.001}$ \\
\cmidrule{1-10} \cmidrule{2-10}
\multirow[c]{2}{*}{\texttt{rel-event}} & \multirow[c]{2}{*}{\texttt{user-attendance}} & Val & $-0.249$ & $\bm{-0.037}$ & $-0.249$ & $-0.193$ & $-0.147$ & $-0.249$ & $-0.045$ \\
 &  & Test & $-0.168$ & $-0.019$ & $-0.168$ & $-0.065$ & $-0.043$ & $-0.168$ & $\bm{0.003}$ \\
\cmidrule{1-10} \cmidrule{2-10}
\multirow[c]{2}{*}{\texttt{rel-f1}} & \multirow[c]{2}{*}{\texttt{driver-position}} & Val & $-5.715$ & $-0.370$ & $-0.236$ & $-2.866$ & $-2.840$ & $0.150$ & $\bm{0.249}$ \\
 &  & Test & $-5.239$ & $-0.119$ & $-0.042$ & $-2.841$ & $-2.849$ & $\bm{0.068}$ & $0.039$ \\
\cmidrule{1-10} \cmidrule{2-10}
\multirow[c]{2}{*}{\texttt{rel-hm}} & \multirow[c]{2}{*}{\texttt{item-sales}} & Val & $-0.017$ & $-0.000$ & $-0.017$ & $0.065$ & $0.053$ & $-0.017$ & $\bm{0.187}$ \\
 &  & Test & $-0.017$ & $-0.000$ & $-0.017$ & $0.058$ & $0.042$ & $-0.017$ & $\bm{0.215}$ \\
\cmidrule{1-10} \cmidrule{2-10}
\multirow[c]{2}{*}{\texttt{rel-ratebeer}} & \multirow[c]{2}{*}{\texttt{user-count}} & Val & $-0.037$ & $-0.053$ & $-0.037$ & $0.551$ & $0.547$ & $\bm{0.559}$ & $0.526$ \\
 &  & Test & $-0.071$ & $-0.025$ & $-0.071$ & $0.264$ & $0.285$ & $-0.170$ & $\bm{0.625}$ \\
\cmidrule{1-10} \cmidrule{2-10}
\multirow[c]{2}{*}{\texttt{rel-stack}} & \multirow[c]{2}{*}{\texttt{post-votes}} & Val & $-0.028$ & $-0.007$ & $-0.028$ & $\bm{0.306}$ & $0.285$ & $-0.028$ & $0.122$ \\
 &  & Test & $-0.034$ & $-0.004$ & $-0.034$ & $\bm{0.294}$ & $0.272$ & $-0.034$ & $0.122$ \\
\cmidrule{1-10} \cmidrule{2-10}
\multirow[c]{4}{*}{\texttt{rel-trial}} & \multirow[c]{2}{*}{\texttt{site-success}} & Val & $-0.988$ & $\bm{-0.005}$ & $-0.988$ & $-0.749$ & $-0.809$ & $-0.319$ & $-0.425$ \\
 &  & Test & $-0.923$ & $\bm{-0.001}$ & $-0.923$ & $-0.714$ & $-0.751$ & $-0.336$ & $-0.483$ \\
\cmidrule{2-10}
 & \multirow[c]{2}{*}{\texttt{study-adverse}} & Val & $-0.021$ & $-0.002$ & $-0.020$ & $-0.021$ & $-0.021$ & $\bm{0.134}$ & $0.066$ \\
 &  & Test & $-0.054$ & $-0.005$ & $-0.050$ & $-0.054$ & $-0.054$ & $\bm{0.307}$ & $0.177$ \\
\bottomrule
\end{tabular}

%% file: tables/results/recommendation_main.tex
\begin{tabular}{lllrrrrrrr}
\toprule
Dataset & Task & Split & Past Visit & Global Pop. & LightGBM & GNN (2) & GNN (4) & IDGNN (2) & IDGNN (4) \\
\midrule
\multirow[c]{2}{*}{\texttt{rel-arxiv}} & \multirow[c]{2}{*}{\texttt{paper-paper-cocitation}} & Val & $19.01$ & $1.25$ & $12.49$ & $12.19$ & $12.96$ & $25.22$ & $\bm{35.76}$ \\
 &  & Test & $16.51$ & $1.13$ & $11.01$ & $8.83$ & $10.46$ & $22.95$ & $\bm{35.39}$ \\
\cmidrule{1-10} \cmidrule{2-10}
\multirow[c]{2}{*}{\texttt{rel-f1}} & \multirow[c]{2}{*}{\texttt{driver-circuit-compete}} & Val & $53.41$ & $55.19$ & $66.06$ & $3.60$ & $10.57$ & $70.70$ & $\bm{74.40}$ \\
 &  & Test & $20.76$ & $50.12$ & $57.77$ & $9.67$ & $16.57$ & $62.32$ & $\bm{76.18}$ \\
\cmidrule{1-10} \cmidrule{2-10}
\multirow[c]{6}{*}{\texttt{rel-ratebeer}} & \multirow[c]{2}{*}{\texttt{user-beer-favorite}} & Val & $0.00$ & $2.33$ & $1.24$ & $2.09$ & $-$ & $3.09$ & $\bm{3.33}$ \\
 &  & Test & $0.00$ & $1.10$ & $0.67$ & $0.56$ & $-$ & $1.21$ & $\bm{1.89}$ \\
\cmidrule{2-10}
 & \multirow[c]{2}{*}{\texttt{user-beer-liked}} & Val & $0.00$ & $0.77$ & $0.43$ & $0.77$ & $—$ & $0.21$ & $\bm{1.48}$ \\
 &  & Test & $0.00$ & $0.61$ & $0.29$ & $0.54$ & $-$ & $0.32$ & $\bm{1.46}$ \\
\cmidrule{2-10}
 & \multirow[c]{2}{*}{\texttt{user-place-liked}} & Val & $0.00$ & $0.24$ & $0.24$ & $1.06$ & $—$ & $0.88$ & $\bm{2.20}$ \\
 &  & Test & $0.00$ & $0.11$ & $0.08$ & $1.15$ & $-$ & $0.60$ & $\bm{1.85}$ \\
\cmidrule{1-10} \cmidrule{2-10}
\multicolumn{2}{c}{\multirow{2}{*}{Average}} & Val & $14.48$ & $11.96$ & $16.09$ & $3.94$ & $11.76$ & $20.02$ & $\bm{23.43}$ \\
 &  & Test & $7.45$ & $10.61$ & $13.96$ & $4.15$ & $13.52$ & $17.48$ & $\bm{23.35}$ \\
\bottomrule
\end{tabular}

%% file: sections/appendix-dataset-schemas.tex
\begin{figure}[!ht]
    \centering
    \includegraphics[width=0.75\linewidth]{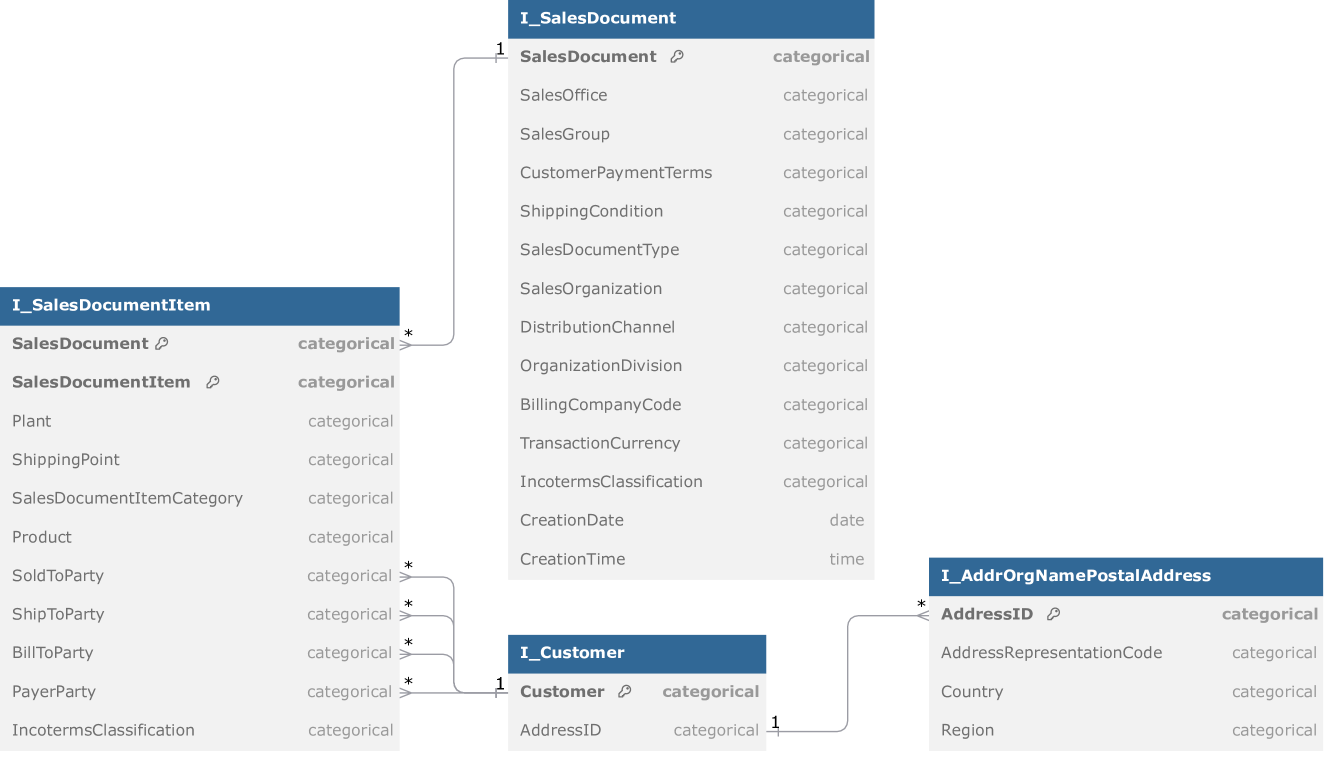}
    \caption{\relbench schema of the newly added Sales Autocompletion Linked Business Tables (SALT) dataset~\citep{sap-salt}.}
    \label{fig:salt-schema}
\end{figure}

\begin{figure}[!ht]
    \centering
    \includegraphics[width=0.75\linewidth]{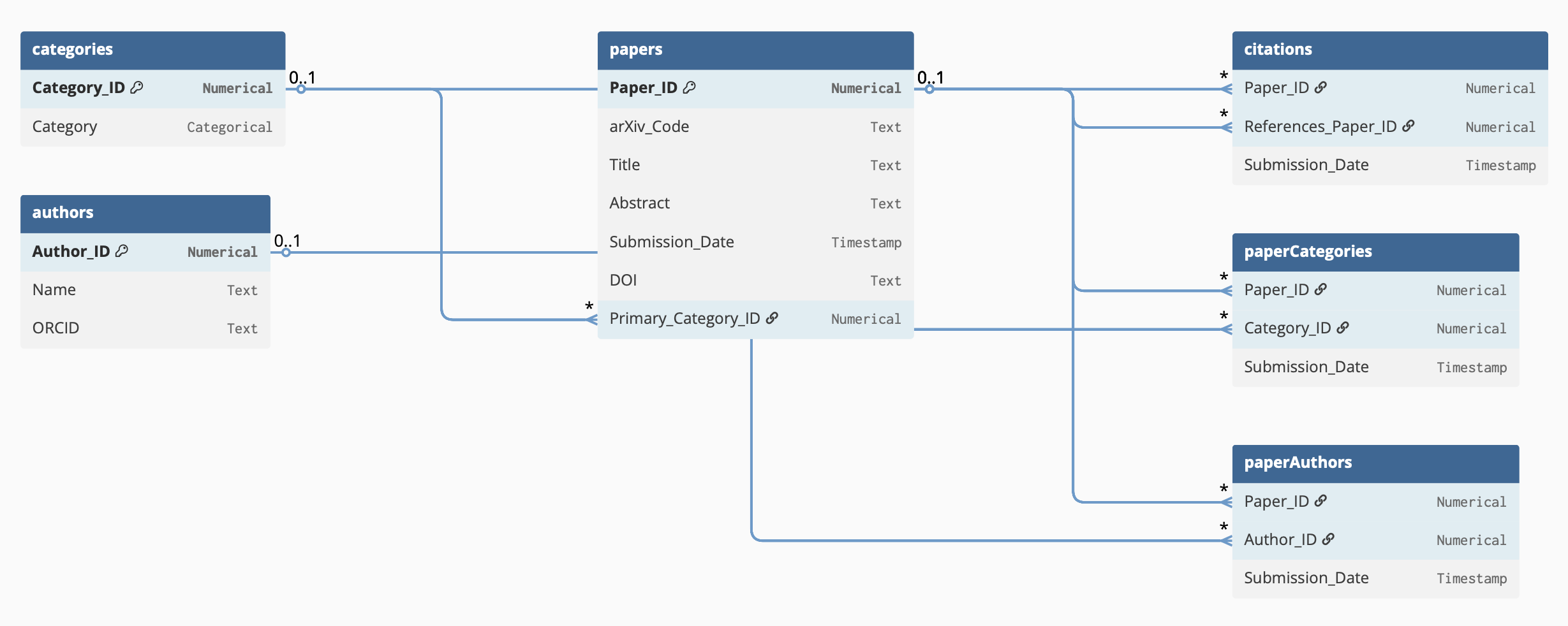}
    \caption{\relbench schema of the newly added arXiv-physics dataset~\citep{arxiv_physics_dataset}.}
    \label{fig:arxiv-schema}
\end{figure}

\begin{figure}[!ht]
    \centering
    \includegraphics[width=0.7\linewidth]{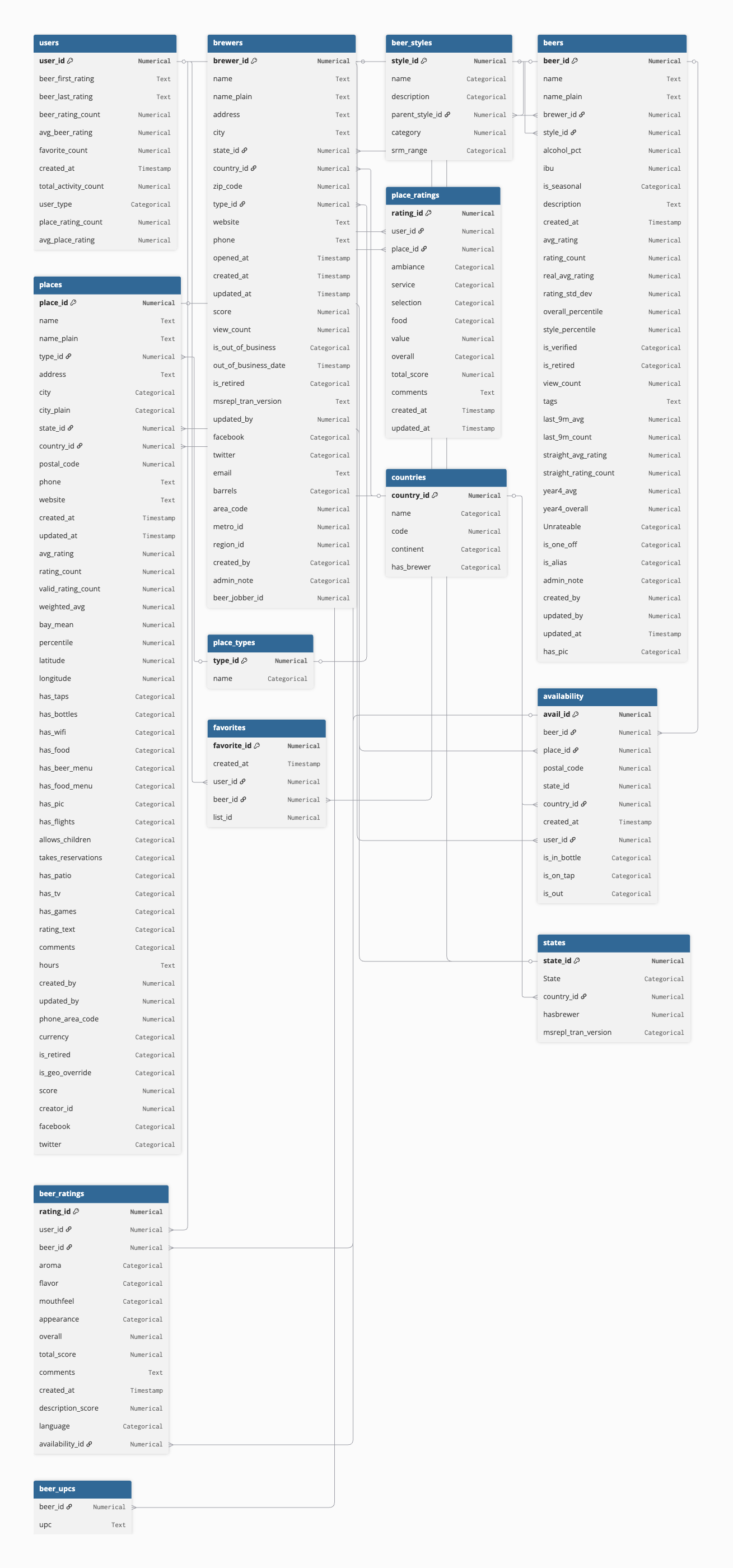}
    \caption{\relbench schema of the newly added RateBeer dataset.}
    \label{fig:ratebeer-schema}
\end{figure}

\begin{figure}[!ht]
    \centering
    \includegraphics[width=0.7\linewidth]{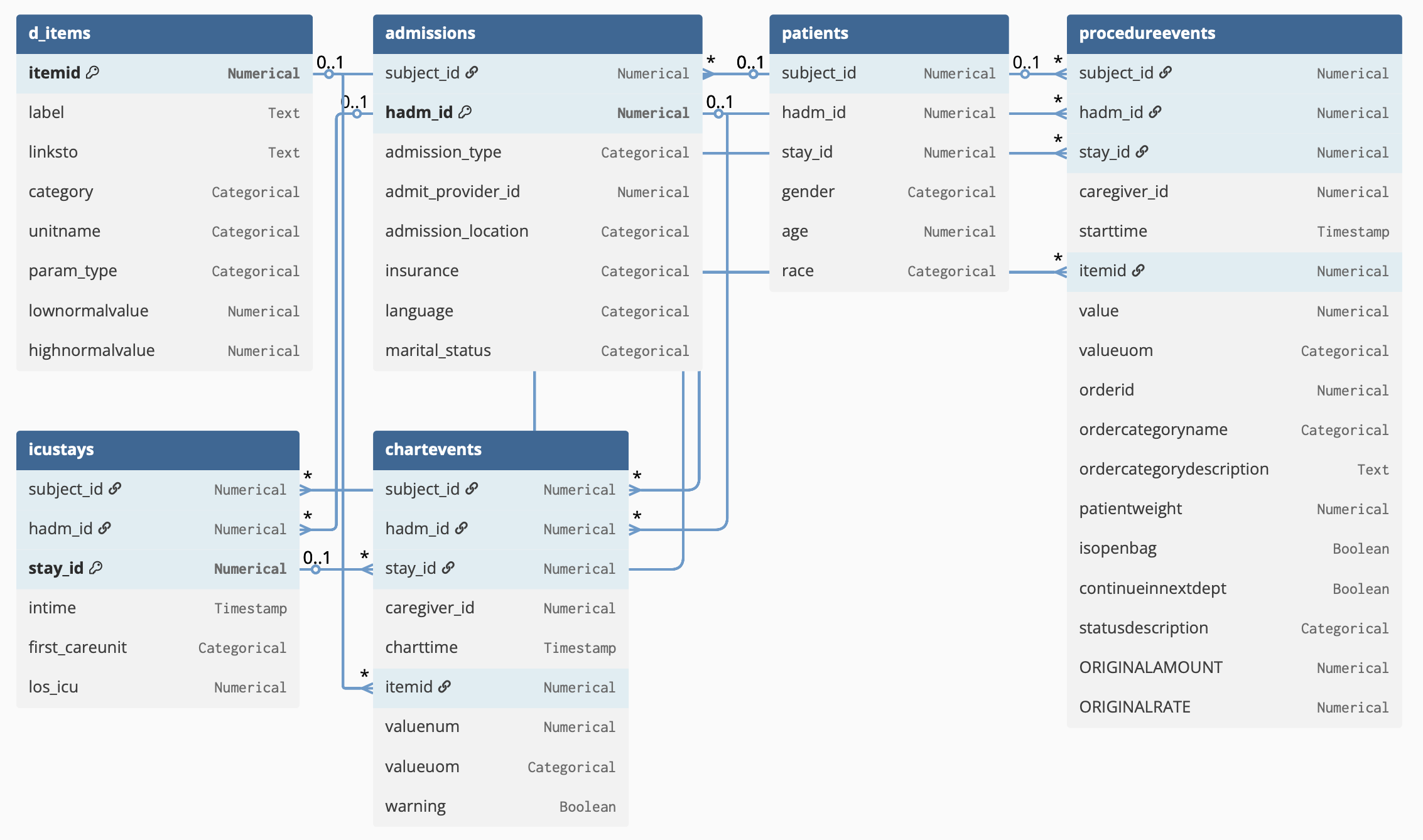}
    \caption{\relbench schema of the newly added MIMIC-IV v3.1 dataset~\citep{johnson2024mimic}.}
    \label{fig:ratebeer-schema}
\end{figure}

%% file: sections/appendix-task-descriptions.tex
\subsection{Autocomplete task: motivation}

\begin{figure}[!ht]
    \centering
    \includegraphics[width=0.8\linewidth]{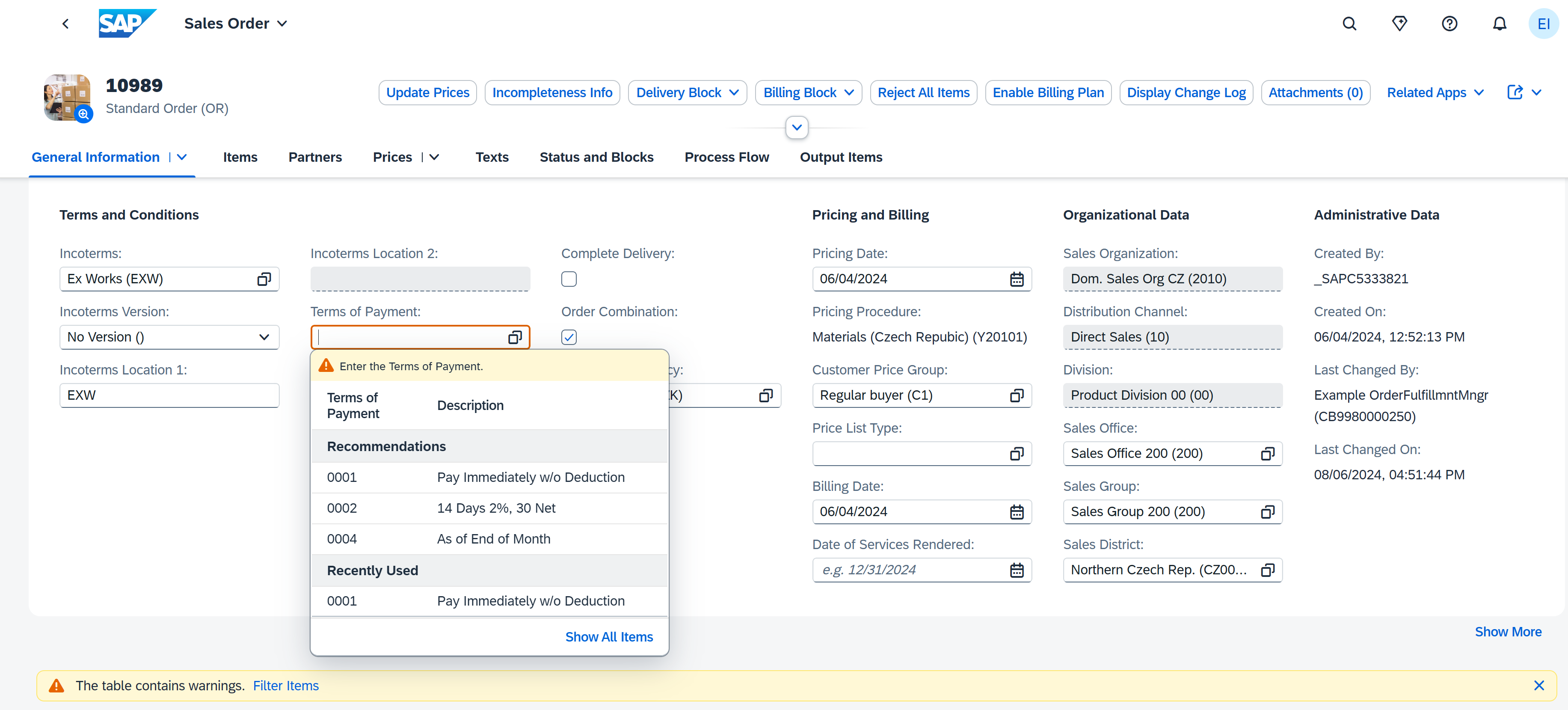}
    \caption{Illustrative example of a real-world autocomplete task, where the SAP S/4HANA Sales Order User interface \citep{sap-salt} predicts payment terms based on other filled-in response fields.}
    \label{fig:autocomplete-example}
\end{figure}

Autocomplete tasks were inspired by the sales order autocomplete task from the SAP S/4HANA Sales
Order User interface. In Figure \ref{fig:autocomplete-example}, the response fields as a whole correspond to one row in a dataset. The user has filled in most of the response fields, allowing the interface to predict the terms of payment for this record.

\subsection{List of predictive task descriptions}

\begin{enumerate}
  \item \texttt{rel-amazon}
    
    Autocomplete Regression:
    \begin{itemize}
      \item \texttt{review-rating}: For each review, predict the star rating.
    \end{itemize}

  \item \texttt{rel-arxiv}

  Forecasting Classification:
  \begin{itemize}
    \item \texttt{paper-citation}: For each paper, predict whether it will receive at least one citation in the next 6 months.
    \item \texttt{author-category}: For each author, predict the primary research category in which they will publish most in the next 6 months.
  \end{itemize}

  Forecasting Regression:
  \begin{itemize}
    \item \texttt{author-publication}: For each author, predict how many papers they will publish in the next 6 months.
  \end{itemize}

  Recommendation:
  \begin{itemize}
    \item \texttt{paper-paper-cocitation}: For each paper, predict which other papers will be co-cited with it in the next 6 months.
  \end{itemize}

  \item \texttt{rel-avito}

  Autocomplete Classification:
  \begin{itemize}
    \item \texttt{searchstream-click}: For each search session, predict whether the user clicked on a result.
    \item \texttt{searchinfo-isuserloggedon}: For each search, predict whether the user was logged in.
  \end{itemize}

  \item \texttt{rel-trial}

  Autocomplete Classification:
  \begin{itemize}
    \item \texttt{studies-has\_dmc}: For each study, predict whether it has a data monitoring committee.
    \item \texttt{eligibilities-adult}: For each study, predict whether it includes adult participants.
    \item \texttt{eligibilities-child}: For each study, predict whether it includes child participants.
  \end{itemize}

  Autocomplete Regression:
  \begin{itemize}
    \item \texttt{studies-enrollment}: For each study, predict the enrollment count.
  \end{itemize}

  \item \texttt{rel-event}

  Autocomplete Classification:
  \begin{itemize}
    \item \texttt{event\_interest-interested}: For each user--event interaction, predict whether the user marked the event as ``interested.''
    \item \texttt{event\_interest-not\_interested}: For each user--event interaction, predict whether the user marked the event as ``not interested.''
  \end{itemize}

  Autocomplete Regression:
  \begin{itemize}
    \item \texttt{users-birthyear}: For each user, predict the user's birth year.
  \end{itemize}

  \item \texttt{rel-f1}

  Autocomplete Regression:
  \begin{itemize}
    \item \texttt{results-position}: For each race result, predict the finishing position.
    \item \texttt{qualifying-position}: For each qualifying entry, predict the qualifying position.
  \end{itemize}

  Recommendation:
  \begin{itemize}
    \item \texttt{driver-circuit-compete}: Predict in which circuits a driver will compete in the next year.
  \end{itemize}

  \item \texttt{rel-hm} 

  Autocomplete Regression:
  \begin{itemize}
    \item \texttt{transactions-price}: For each transaction, predict the item price.
  \end{itemize}

  \item \texttt{rel-ratebeer} 

  Autocomplete Regression:
  \begin{itemize}
    \item \texttt{beer\_ratings-total\_score}: For each user, given a beer, predict the total score rating the user will give to the beer.
  \end{itemize}

  Forecasting Classification:
  \begin{itemize}
    \item \texttt{beer-churn}: For each beer, predict if it will receive zero ratings in the next 90 days.
    \item \texttt{user-churn}: For each active user, predict if they will rate zero beers in the next 90 days.
    \item \texttt{brewer-dormant}: For each brewer, predict if it will release zero beers in the next year (risk of going dormant).
  \end{itemize}

  Forecasting Regression:
  \begin{itemize}
    \item \texttt{user-count}: Predict the number of ratings a user will give in the next 90 days.
  \end{itemize}

  Recommendation:
  \begin{itemize}
    \item \texttt{user-beer-favorite}: For each user, predict the top 10 beers they will next add to their Favorites list.
    \item \texttt{user-beer-liked}: For each user, predict the top 10 beers they will rate at least 4.0 / 5.0.
    \item \texttt{user-place-liked}: For each user, predict the top 10 places they will rate at least 80 / 100.
  \end{itemize}

  \item \texttt{rel-salt}

  Autocomplete Classification:
  \begin{itemize}
    \item \texttt{item-plant}: For each sales order item, predict its plant (production/storage facility).
    \item \texttt{item-shippoint}: For each sales order item, predict its shipping point (dispatch location).
    \item \texttt{item-incoterms}: For each sales order item, predict its item-level international commercial terms.
    \item \texttt{sales-office}: For each sales order, predict the sales office responsible for managing sales activities for the relevant products and geographic region.
    \item \texttt{sales-group}: For each sales order, predict the sales group, i.e. the subdivision within the distribution chain that handles the customer and transaction.
    \item \texttt{sales-payterms}: For each sales order, predict the customer payment terms (payment deadlines/discounts).
    \item \texttt{sales-shipcond}: For each sales order, predict the shipping condition (logistics terms).
    \item \texttt{sales-incoterms}: Predict the header-level Incoterms (international commercial terms) for each sales order.
  \end{itemize}

  \item \texttt{rel-stack} 

  Autocomplete Classification:
  \begin{itemize}
    \item \texttt{badges-class}: For each badge, predict the badge class.
  \end{itemize}

  \item \texttt{rel-mimic} 

  Forecasting Classification:
  \begin{itemize}
    \item \texttt{patient-iculengthofstay}: For each patient admitted into the ICU, predict whether their stay will last at least 3 days.
  \end{itemize}

\end{enumerate}

%% file: sections/appendix-task-std.tex
\subsection{Results with standard deviations}

Tables \ref{tab:autocomplete_classification_appendix} - \ref{tab:recommendation_appendix} show mean and standard deviations over 5 runs for the autocomplete classification, autocomplete regression, entity classification, entity regression and link prediction results for all new tasks introduced in \relbench v2. For regression tasks, MAE metrics are reported below (the main paper reports $R^2$ metrics).

\begin{table}[!ht]
  \centering
  \caption{{\bf Autocomplete binary classification results on \relbench.} Binary classification uses the AUROC metric (higher is better). Standard baselines of random choice and majority class both correspond to AUROC values of approximately 50.00, so we exclude them below. Best values are in bold.}
  \label{tab:autocomplete_classification_appendix}
  \renewcommand{\arraystretch}{1.1}
  \scriptsize
  \resizebox{\textwidth}{!}{
    \input{tables/results/autocomplete_classification_appendix}
  }
  \vspace{-10pt}
\end{table}

\begin{table}[!ht]
  \centering
  \caption{{\bf Autocomplete multiclass classification results on \relbench.} Multiclass classification uses the accuracy metric (higher is better). Best values are in bold.}
  \label{tab:autocomplete_multiclass_appendix}
  \renewcommand{\arraystretch}{1.1}
  \scriptsize
    \input{tables/results/autocomplete_multiclass_appendix}
  \vspace{-10pt}
\end{table}

\begin{table}[!ht]
  \centering
  \caption{{\bf Autocomplete regression $R^2$ results on \relbench.} Regression uses the $R^2$ metric (higher is better). Best values are in bold.}
  \label{tab:autocomplete_regression_r2_appendix}
  \renewcommand{\arraystretch}{1.1}
  \scriptsize
    \resizebox{\textwidth}{!}{
    \input{tables/results/autocomplete_regression_r2_appendix}
    }
  \vspace{10pt}
\end{table}

\begin{table}[!ht]
  \centering
  \caption{{\bf Autocomplete regression results on \relbench.} Regression uses the MAE metric (lower is better). Best values are in bold.}
  \label{tab:autocomplete_regression_appendix}
  \renewcommand{\arraystretch}{1.1}
  \scriptsize
    \resizebox{\textwidth}{!}{
    \input{tables/results/autocomplete_regression_appendix}
    }
  \vspace{10pt}
\end{table}

\begin{table}[!ht]
  \centering
  \caption{{\bf Entity binary classification results on \relbench.} Binary classification uses the AUROC metric (higher is better). Standard baselines of random choice and majority class both correspond to AUROC values of approximately 50.00, so we exclude them below. Best values are in bold.}
  \label{tab:entity_classification_appendix}
  \renewcommand{\arraystretch}{1.1}
  \scriptsize
    \input{tables/results/entity_classification_appendix}
  \vspace{10pt}
\end{table}

\begin{table}[!ht]
  \centering
  \caption{{\bf Entity multiclass classification results on \relbench.} Multiclass classification uses the accuracy metric (higher is better). Best values are in bold.}
  \label{tab:entity_multiclass_appendix}
  \renewcommand{\arraystretch}{1.1}
  \scriptsize
    \input{tables/results/entity_multiclass_appendix}
  \vspace{10pt}
\end{table}

\begin{table}[!ht]
  \centering
  \caption{{\bf Entity regression $R^2$ results on \relbench.} Regression uses the $R^2$ metric (higher is better). Best values are in bold.}
  \label{tab:entity_regression_r2_appendix}
  \renewcommand{\arraystretch}{1.1}
  \scriptsize
    \resizebox{\textwidth}{!}{
    \input{tables/results/entity_regression_r2_appendix}
    }
  \vspace{10pt}
\end{table}

\begin{table}[!ht]
  \centering
  \caption{{\bf Entity regression results on \relbench.} Regression uses the MAE metric (lower is better). Best values are in bold.}
  \label{tab:entity_regression_appendix}
  \renewcommand{\arraystretch}{1.1}
  \scriptsize
    \resizebox{\textwidth}{!}{
    \input{tables/results/entity_regression_appendix}
    }
  \vspace{10pt}
\end{table}

\begin{table}[!ht]
  \centering
  \caption{{\bf Recommendation results on \relbench.} Recommendation uses the MAP metric (higher is better). Best values are in bold. Results for 4-layer GNN and ID-GNN models on \relbench v1 recommendation tasks are also included below.}
  \label{tab:recommendation_appendix}
  \renewcommand{\arraystretch}{1.1}
  \scriptsize
    \resizebox{\textwidth}{!}{
    \input{tables/results/recommendation_appendix}
    }
  \vspace{10pt}
\end{table}

\subsection{Additional regression metrics}

For regression tasks, the main paper reports $R^2$ to measure each method's predictive power, while MAE is reported in Tables \ref{tab:autocomplete_regression_appendix} and \ref{tab:entity_regression_appendix} above.

\newpage

\subsection{Recommendation task ablations}

\textbf{The benefit of node position}. ID-GNN, which strongly utilizes node position, excels in entity-specific tasks. For example, we focus on the \ratebeer dataset whose recommendation tasks strongly highlight user-specific preferences. We observed that ID-GNN significantly outperforms plain GraphSAGE while also being substantially faster to train. This advantage stems from the fact that recommendation tasks are inherently multi-hop link prediction problems, where node position and context are crucial, as nodes tend to connect with others in their community. ID-GNN is better at capturing such node-specific patterns than vanilla GraphSAGE, leading to its superior performance.

\textbf{Multi-hop recommendation}. Recommendation tasks naturally involve multi-hop reasoning, where predicting a user–item link often requires traversing intermediate tables in the relational schema. Increasing the number of GNN layers expands the receptive field, allowing the model to aggregate information from progressively richer neighborhoods. This is particularly beneficial when intermediate tables contain strong preference signals. In the \texttt{user-beer-liked} task, the Beer Ratings table includes explicit ratings, subscores, and review text, enabling deeper GNNs to capture collaborative filtering effects by traversing paths such as User → Ratings → Beer → Ratings → Beer. In such cases, increasing the number of layers yields substantial performance gains. In contrast, tasks with feature-sparse intermediate tables (e.g., Favorites, which contains the implicit link between users and beers with timestamps) tend to benefit less from additional hops, as deeper neighborhoods introduce weaker or noisier signals.

\subsection{Experiment hyperparameters}

All experiments for \relbench v2 were conducted with minimal parameter tuning. The default
hyperparameters are specified in Table~\ref{tab:hparams}. The consistency of these defaults across
task types highlights the robustness of RDL models, which perform well without extensive
hyperparameter optimization.

For the \texttt{rel-ratebeer} recommendation tasks, we adjusted the batch size to 64 when training
two-layer GraphSAGE, two-layer ID-GNN, and four-layer ID-GNN models to accommodate GPU memory
constraints. While four-layer GraphSAGE remains prohibitively memory-intensive under this setting,
all other models were trained successfully with the adjusted configuration.

\begin{table}[t]
  \centering
  \caption{{Task-specific RDL default hyperparameters.}}
  \label{tab:hparams}
  \renewcommand{\arraystretch}{1.1}
  \setlength{\tabcolsep}{3pt}
  \scriptsize
  \begin{tabular}{lrrrr}
    \toprule
      \mr{2}{\textbf{Hyperparameter}} & \mc{4}{c}{\textbf{Task type}} \\
      & Autocomplete & Entity classification & Entity regression & Recommendation \\
    \midrule
      Learning rate & 0.005 & 0.005 & 0.005 & 0.001 \\
      Maximum epochs & 10 & 10 & 10 & 20 \\
      Batch size & 512 & 512 & 512 & 512 \\
      Hidden feature size & 128 & 128 & 128 & 128 \\
      Aggregation & summation & summation & summation & summation \\
      Number of layers & 2 & 2 & 2 & 2 \\
      Number of neighbors & 128 & 128 & 128 & 128 \\
      Temporal sampling strategy & uniform & uniform & uniform & uniform \\
   \bottomrule
  \end{tabular}
\end{table}

\section{Bridging Temporal Graph Benchmark (TGB) and \relbench} \label{sec:tgb_relbench}

\paragraph{Dataset descriptions.}
\texttt{tgbl-wiki-v2} captures Wikipedia co-edit interactions over a short horizon and is naturally bipartite; \texttt{tgbl-review-v2} is a long-range e-commerce interaction graph derived from Amazon review activity; \texttt{tgbl-coin} consists of cryptocurrency transactions; \texttt{tgbl-comment} models a large-scale Reddit comment interaction stream; and \texttt{tgbl-flight} is a decades-long flight network event log.
For node property prediction, \texttt{tgbn-trade} models annual UN trade interactions, \texttt{tgbn-genre} is a LastFM user--genre interaction stream, \texttt{tgbn-reddit} is a temporal Reddit hyperlink network, and \texttt{tgbn-token} represents blockchain user--token interactions.
Finally, the heterogeneous family \texttt{thgl-*} includes GitHub interaction streams (\texttt{thgl-software}, \texttt{thgl-github}), a Reddit interaction stream (\texttt{thgl-forum}), and an app-market interaction stream (\texttt{thgl-myket}), each with explicit node/edge type constraints.

\subsection{Translation: temporal event streams to relational schemas} \label{sec:tgb_translation}

Each TGB dataset is provided as a chronological stream of interactions. Our translation materializes a \relbench-style normalized schema in which \emph{entities} become tables with primary keys and \emph{events} become tables whose rows reference entities via foreign keys and include a time column. Across all translated datasets, we (i) include an explicit timestamp column (\texttt{event\_ts}) on time-varying tables; (ii) avoid storing train/validation/test split columns, and instead store dataset-level cutoffs (\valtimestamp, \testtimestamp) as metadata so splits can be derived from timestamps at load/evaluation time; and (iii) keep only low-dimensional attributes as relational columns (e.g., scalar \texttt{weight}), while omitting high-dimensional edge message vectors that would otherwise expand into hundreds of sparse columns.

For dynamic link prediction datasets (\texttt{tgbl-*}), we represent the interaction stream as a single event table referencing a node table:
\begin{quote}
  \small
  \texttt{nodes(node\_id)}\\
  \texttt{events(event\_id, src\_id, dst\_id, event\_ts, weight)}
\end{quote}
\noindent For bipartite datasets (e.g., \texttt{tgbl-wiki-v2}), we use separate entity tables \texttt{src\_nodes(src\_id)} and \texttt{dst\_nodes(dst\_id)} to preserve type constraints and enable type-correct negative sampling.

For temporal heterogeneous datasets (\texttt{thgl-*}), we materialize one entity table per node type and one event table per edge type:
\begin{quote}
  \small
  \texttt{nodes\_type\_t(node\_type\_t\_id)} for each node type \(t\).\\
  \texttt{events\_edge\_type\_e(event\_id, src\_id, dst\_id,}\\
  \texttt{event\_ts, weight)} for each edge type \(e\).
\end{quote}
\noindent This schema-first representation preserves the semantics of typed relations and makes relation-conditioned tasks and evaluation natural in \relbench.

For dynamic node property prediction datasets (\texttt{tgbn-*}), targets are time-varying and often sparse. To avoid storing dense label vectors, we represent supervision as normalized label events:
\begin{quote}
  \small
  \texttt{labels(label\_id)}\\
  \texttt{label\_events(label\_event\_id, src\_id, label\_ts)}\\
  \texttt{label\_event\_items(item\_id, label\_event\_id, label\_id,}\\
  \texttt{label\_weight)}
\end{quote}
\noindent This preserves the full supervision signal while keeping storage proportional to the number of non-zeros.

\subsection{Efficient storage and loading: parquet + disk-backed CSR} \label{sec:tgb_parquet_csr}

  A central engineering challenge in translating TGB to \relbench is scale: several datasets contain on the order of \(\mathcal{O}(10^7\text{--}10^8)\) temporal events. Na\"ively materializing the full event log as an in-memory edge list (e.g., a global
  \texttt{edge\_index}) is prohibitive on commodity CPU machines, so we rely on columnar storage (parquet) together with disk-backed sparse graph caches. We store each table as a separate parquet file in the standard \relbench \texttt{Database.save()} layout. This provides columnar storage for fast projection to only the columns required by a given model, and row-group metadata that supports efficient
  sequential scans without loading entire tables into memory. For \texttt{thgl-*} datasets with many edge types, we additionally export each \texttt{events\_edge\_type\_*} table using a streaming/chunked parquet writer. This avoids materializing massive
  intermediate dataframes per relation and keeps the conversion pipeline memory-bounded even when the total number of events is large.

  To train sampled GNN baselines at scale, we build and cache CSR (compressed sparse row) adjacency representations directly from parquet scans. Concretely, we store \texttt{indptr} and \texttt{indices} arrays (and, when needed, aligned per-event arrays
  such as timestamps and weights) as memory-mappable \texttt{.npy} artifacts. With CSR, neighbor sampling reduces to lightweight slicing operations over these arrays, enabling mini-batch training without ever holding the full graph in system memory. For relational baselines, we use an ``event-as-node'' message passing graph: each event row becomes a node connected (via PK/FK edges) to its incident entity rows. We build node$\rightarrow$event CSR adjacencies per entity table and unify per-event arrays. This makes relational message passing scalable while remaining faithful to the normalized schema.

\begin{algorithm}[t]
\caption{Streaming CSR construction from parquet event logs (sketch).}
\label{alg:csr_from_parquet}
\begin{algorithmic}[1]
\Require Parquet event table with columns \texttt{src\_id}, \texttt{dst\_id}, \texttt{event\_ts}; cutoff time $\tau$.
\Ensure CSR adjacency arrays \texttt{indptr}, \texttt{indices} for events with \texttt{event\_ts} $\le \tau$.
\State \textbf{Pass 1:} Scan parquet in large row batches; filter rows by \texttt{event\_ts} $\le \tau$; accumulate per-node degree counts.
\State Build \texttt{indptr} by prefix-summing degrees.
\State \textbf{Pass 2:} Re-scan parquet in batches; filter by \texttt{event\_ts} $\le \tau$; write neighbors into \texttt{indices} using \texttt{indptr} offsets.
\State Optionally: symmetrize for undirected message passing; cache per-event arrays (\texttt{event\_ts}, \texttt{weight}) aligned with \texttt{indices}.
\end{algorithmic}
\end{algorithm}

\subsection{Baselines and evaluation protocol} \label{sec:tgb_baselines}

  We benchmark three complementary model families on the translated TGB datasets, spanning graph-native GNNs, relational (PK/FK) GNNs, and temporal sequence models.

  \textbf{GraphSAGE (projected-edge graph; TGB-style).}
  We treat each interaction event as an edge from \texttt{src} to \texttt{dst} and train a sampled two-layer GraphSAGE encoder using disk-backed CSR adjacency built from parquet scans. This baseline is graph-native in the sense that the event log is
  directly interpreted as a temporal edge stream over a single (or bipartite) node set.

  \textbf{GraphSAGE (event-as-node relational graph; RDL-style).}
  We represent each event row as its own node and perform message passing over the induced PK/FK graph that connects entity rows to event rows (``RelEventSAGE''). This baseline is schema-faithful: neighborhoods and aggregation follow the relational
  structure rather than collapsing the database into a single projected graph.

  \textbf{TGN + attention (temporal baseline).}
  We train a temporal graph network with a lightweight attention-based embedding module that consumes the chronological event stream (or equivalently, the exported parquet event tables) under controlled compute budgets. This baseline captures temporal
  dynamics via memory and time encoding, and is directly comparable to the GNN baselines under the same split and leakage constraints.

  \textbf{Metrics.}
  For \texttt{tgbl-*} and \texttt{thgl-*} we report sampled-negative MRR@100 (higher is better) to keep a single, consistent ranking metric across model families and schema variants. For \texttt{tgbn-*} we report the official NDCG@10 (higher is better),
  and additionally report sampled-negative MRR for consistency across task families. For existing \relbench recommendation tasks we report the official MAP@10 used in the \relbench evaluation code.

  \textbf{Leakage control.}
  All splits are derived from dataset-level \valtimestamp and \testtimestamp cutoffs, and we do not store split membership as a database column. When comparing baselines, we ensure that message passing and neighborhood construction only use historical
  events up to the validation cutoff (i.e., \texttt{adj=val}) unless explicitly stated otherwise, preventing test evaluation from incorporating post-cutoff interactions.

\subsection{Results} \label{sec:tgb_results}

Tables \ref{tab:tgbn_nodeprop_results}, \ref{tab:tgbl_linkpred_results}, and \ref{tab:thgl_linkpred_results} summarize baseline performance on the translated TGB datasets. Table \ref{tab:relbench_reco_smoke} reports a small reference point on existing \relbench recommendation tasks using GraphSAGE and the attention-based TGN baseline.

  \begin{table}[!ht]                                                                                                                                                                                                                                         
    \centering                                                                                                                                                                                                                                               
    \caption{{\bf Dynamic node property prediction (\texttt{tgbn-*}) on translated TGB datasets.}                                                                                                                                                            
    We report validation and test performance for GraphSAGE and a sampled neighbor-attention variant (``Attn''), using NDCG@10 (official) and sampled-negative MRR.                                                                                          
    }                                                                                                                                                                                                                                                        
    \label{tab:tgbn_nodeprop_results}                                                                                                                                                                                                                        
    \begingroup                                                                                                                                                                                                                                              
    \scriptsize                                                                                                                                                                                                                                              
    \setlength{\tabcolsep}{2.5pt}                                                                                                                                                                                                                            
    \renewcommand{\arraystretch}{1.1}                                                                                                                                                                                                                        
    \begin{tabular}{lrrrrrrrr}                                                                                                                                                                                                                               
      \toprule                                                                                                                                                                                                                                               
        & \mc{4}{c}{GraphSAGE} & \mc{4}{c}{Attn (sampled neighbor attention)} \\                                                                                                                                                                             
        \cmidrule(lr){2-5} \cmidrule(lr){6-9}                                                                                                                                                                                                                
      Dataset                                                                                                                                                                                                                                                
        & Val MRR & Val NDCG@10 & Test MRR & Test NDCG@10                                                                                                                                                                                                    
        & Val MRR & Val NDCG@10 & Test MRR & Test NDCG@10 \\                                                                                                                                                                                                 
      \midrule                                                                                                                                                                                                                                               
      \texttt{tgbn-trade}                                                                                                                                                                                                                                    
        & 0.9522 & \textbf{0.3769} & 0.9393 & \textbf{0.3765}                                                                                                                                                                                                
        & 0.9102 & 0.3849 & 0.8635 & 0.3401 \\                                                                                                                                                                                                               
      \texttt{tgbn-genre}                                                                                                                                                                                                                                    
        & \textbf{0.8682} & \textbf{0.6169} & \textbf{0.8591} & \textbf{0.6054}                                                                                                                                                                              
        & 0.6664 & 0.4263 & 0.6552 & 0.4139 \\                                                                                                                                                                                                               
      \texttt{tgbn-reddit}                                                                                                                                                                                                                                   
        & \textbf{0.7804} & \textbf{0.6474} & \textbf{0.7555} & \textbf{0.6146}                                                                                                                                                                              
        & 0.4326 & 0.3413 & 0.4067 & 0.3098 \\                                                                                                                                                                                                               
      \texttt{tgbn-token}                                                                                                                                                                                                                                    
        & \textbf{0.4043} & \textbf{0.3763} & \textbf{0.3405} & \textbf{0.3098}                                                                                                                                                                              
        & 0.2451 & 0.2303 & 0.2101 & 0.1935 \\                                                                                                                                                                                                               
      \bottomrule                                                                                                                                                                                                                                            
    \end{tabular}
    \endgroup
    \vspace{-5pt}
  \end{table}

  \begin{table}[!ht]
    \centering
    \caption{{\bf Dynamic link prediction (\texttt{tgbl-*}) on translated TGB datasets (sampled-negative MRR@100).}
    We compare (i) a graph-native projected-edge GraphSAGE baseline, (ii) an event-as-node relational GraphSAGE baseline (RelEventSAGE), and (iii) a TGN + attention baseline trained from exported parquet under bounded budgets.
    Best test results per dataset are in bold.
    }
    \label{tab:tgbl_linkpred_results}
    \begingroup
    \scriptsize
    \setlength{\tabcolsep}{3pt}
    \renewcommand{\arraystretch}{1.1}
    \begin{tabular}{lrrrrrr}
      \toprule
        & \mc{2}{c}{GraphSAGE (projected edges)} & \mc{2}{c}{GraphSAGE (event-as-node)} & \mc{2}{c}{TGN+Attn} \\
        \cmidrule(lr){2-3} \cmidrule(lr){4-5} \cmidrule(lr){6-7}
      Dataset & Val & Test & Val & Test & Val & Test \\
      \midrule
      \texttt{tgbl-wiki-v2}   & \textbf{0.4203} & \textbf{0.3782} & 0.2757 & 0.2517 & 0.3998 & 0.3384 \\
      \texttt{tgbl-review-v2} & 0.0932 & 0.0852 & 0.2596 & 0.2317 & \textbf{0.2528} & \textbf{0.2457} \\
      \texttt{tgbl-coin}      & 0.4541 & 0.3932 & \textbf{0.6064} & \textbf{0.5554} & 0.5604 & 0.5067 \\
      \texttt{tgbl-comment}   & 0.2089 & 0.1536 & \textbf{0.2896} & \textbf{0.2305} & 0.2960 & 0.2098 \\
      \texttt{tgbl-flight}    & \textbf{0.7082} & \textbf{0.6737} & 0.6357 & 0.5915 & 0.4838 & 0.4566 \\
      \bottomrule
    \end{tabular}
    \endgroup
    \vspace{-5pt}
  \end{table}

  \begin{table}[!ht]
    \centering
    \caption{{\bf Temporal heterogeneous link prediction (\texttt{thgl-*}) on translated TGB datasets (sampled-negative MRR@100).}
    We compare event-as-node relational GraphSAGE (RelEventSAGE) against TGN + attention.
    Best test results per dataset are in bold.
    }
    \label{tab:thgl_linkpred_results}
    \begingroup
    \scriptsize
    \setlength{\tabcolsep}{3pt}
    \renewcommand{\arraystretch}{1.1}
    \begin{tabular}{lrrrr}
      \toprule
        & \mc{2}{c}{GraphSAGE (event-as-node)} & \mc{2}{c}{TGN+Attn} \\
        \cmidrule(lr){2-3} \cmidrule(lr){4-5}
      Dataset & Val & Test & Val & Test \\
      \midrule
      \texttt{thgl-software} & 0.1388 & 0.1206 & 0.1367 & \textbf{0.1290} \\
      \texttt{thgl-forum}    & \textbf{0.4635} & \textbf{0.4401} & 0.3452 & 0.3527 \\
      \texttt{thgl-myket}    & \textbf{0.7264} & \textbf{0.7084} & 0.6614 & 0.6648 \\
      \texttt{thgl-github}   & 0.0725 & 0.0666 & \textbf{0.0782} & \textbf{0.0767} \\
      \bottomrule
    \end{tabular}
    \endgroup
    \vspace{-5pt}
  \end{table}

  \begin{table}[!ht]
    \centering
    \caption{{\bf Reference recommendation results on existing \relbench datasets (smoke runs).}
    We report validation MAP@10 for GraphSAGE and TGN+Attn on three \relbench recommendation tasks.
    }
    \label{tab:relbench_reco_smoke}
    \begingroup
    \scriptsize
    \setlength{\tabcolsep}{3pt}
    \renewcommand{\arraystretch}{1.1}
    \begin{tabular}{lllrr}
      \toprule
      Dataset & Task & Metric & GraphSAGE & TGN+Attn \\
      \midrule
      \fone  & \texttt{driver-race-compete} & val MAP@10 & 0.06048 & \textbf{0.27821} \\
      \texttt{rel-hm} & \texttt{user-item-purchase} & val MAP@10 & 0.0006649 & \textbf{0.0009053} \\
      \stack & \texttt{post-post-related} & val MAP@10 & \textbf{0.0024797} & 0.00017857 \\
      \bottomrule
    \end{tabular}
    \endgroup
    \vspace{-5pt}
  \end{table}

\paragraph{Discussion.}
Across \texttt{tgbl-*}, the comparison between projected-edge GraphSAGE and event-as-node relational GraphSAGE highlights when the relationalization of events is beneficial: datasets with informative event records (e.g., heavy-tailed \texttt{weight} and rich event semantics) tend to favor the event-as-node representation (\texttt{tgbl-coin}, \texttt{tgbl-comment}), while datasets that are closer to ``pure'' structural proximity under our compact schema (and without high-dimensional message features) can favor the projected-edge baseline (\texttt{tgbl-wiki-v2}, \texttt{tgbl-flight}).
On \texttt{thgl-*}, the attention-based temporal baseline is competitive and sometimes best (\texttt{thgl-software}, \texttt{thgl-github}), while relational GraphSAGE remains strong on datasets where type-correct relational neighborhoods provide a high-signal inductive bias (\texttt{thgl-forum}, \texttt{thgl-myket}).
Finally, Table~\ref{tab:relbench_reco_smoke} provides a small anchor on existing \relbench recommendation tasks: even with minimal tuning, the attention-based temporal baseline can substantially improve MAP on some datasets (\fone), but does not uniformly dominate across domains (\stack).

%% file: tables/results/autocomplete_classification_appendix.tex
\begin{tabular}{lllrr}
\toprule
Dataset & Task & Split & LightGBM & GNN \\
\midrule
\multirow[c]{4}{*}{\texttt{rel-avito}} & \multirow[c]{2}{*}{\texttt{searchinfo-isuserloggedon}} & Val & $59.09_{\pm 0.37}$ & $\bm{82.57}_{\pm 0.64}$ \\
 &  & Test & $50.00_{\pm 0.00}$ & $\bm{73.00}_{\pm 0.79}$ \\
\cmidrule{2-5}
 & \multirow[c]{2}{*}{\texttt{searchstream-click}} & Val & $\bm{68.33}_{\pm 0.19}$ & $50.39_{\pm 0.22}$ \\
 &  & Test & $49.92_{\pm 0.17}$ & $\bm{55.92}_{\pm 14.04}$ \\
\cmidrule{1-5} \cmidrule{2-5}
\multirow[c]{4}{*}{\texttt{rel-event}} & \multirow[c]{2}{*}{\texttt{event\_interest-interested}} & Val & $51.25_{\pm 0.00}$ & $\bm{54.16}_{\pm 1.75}$ \\
 &  & Test & $49.57_{\pm 0.00}$ & $47.64_{\pm 3.44}$ \\
\cmidrule{2-5}
 & \multirow[c]{2}{*}{\texttt{event\_interest-not\_interested}} & Val & $51.98_{\pm 0.00}$ & $49.74_{\pm 13.71}$ \\
 &  & Test & $52.88_{\pm 0.00}$ & $\bm{60.40}_{\pm 19.57}$ \\
\cmidrule{1-5} \cmidrule{2-5}
\multirow[c]{6}{*}{\texttt{rel-trial}} & \multirow[c]{2}{*}{\texttt{eligibilities-adult}} & Val & $58.10_{\pm 0.23}$ & $\bm{94.91}_{\pm 0.10}$ \\
 &  & Test & $50.00_{\pm 0.00}$ & $\bm{93.73}_{\pm 0.15}$ \\
\cmidrule{2-5}
 & \multirow[c]{2}{*}{\texttt{eligibilities-child}} & Val & $59.78_{\pm 0.15}$ & $\bm{85.91}_{\pm 0.20}$ \\
 &  & Test & $50.00_{\pm 0.00}$ & $\bm{87.25}_{\pm 0.10}$ \\
\cmidrule{2-5}
 & \multirow[c]{2}{*}{\texttt{studies-has\_dmc}} & Val & $76.47_{\pm 0.26}$ & $\bm{78.21}_{\pm 0.12}$ \\
 &  & Test & $50.00_{\pm 0.00}$ & $\bm{75.72}_{\pm 0.11}$ \\
\bottomrule
\end{tabular}

%% file: tables/results/autocomplete_multiclass_appendix.tex
\begin{tabular}{lllrrrr}
\toprule
Dataset & Task & Split & Random & Majority & LightGBM & GNN \\
\midrule
\multirow[c]{17}{*}{\texttt{rel-salt}} & \multirow[c]{2}{*}{\texttt{item-incoterms}} & Val & $34.49$ & $66.46$ & $66.43_{\pm 0.01}$ & $\bm{80.23}_{\pm 0.48}$ \\
 &  & Test & $30.33$ & $58.05$ & $58.05_{\pm 0.00}$ & $\bm{69.36}_{\pm 0.77}$ \\
\cmidrule{2-7}
 & \multirow[c]{2}{*}{\texttt{item-plant}} & Val & $33.19$ & $60.95$ & $60.97_{\pm 0.04}$ & $\bm{99.70}_{\pm 0.16}$ \\
 &  & Test & $32.38$ & $59.69$ & $59.69_{\pm 0.00}$ & $\bm{99.46}_{\pm 0.12}$ \\
\cmidrule{2-7}
 & \multirow[c]{2}{*}{\texttt{item-shippoint}} & Val & $8.20$ & $2.34$ & $4.72_{\pm 0.02}$ & $\bm{98.54}_{\pm 0.13}$ \\
 &  & Test & $6.53$ & $1.99$ & $5.67_{\pm 5.18}$ & $\bm{98.39}_{\pm 0.08}$ \\
\cmidrule{2-7}
 & \multirow[c]{2}{*}{\texttt{sales-group}} & Val & $0.90$ & $0.86$ & $0.70_{\pm 0.12}$ & $\bm{18.43}_{\pm 0.22}$ \\
 &  & Test & $0.85$ & $0.75$ & $0.94_{\pm 1.32}$ & $\bm{15.76}_{\pm 0.30}$ \\
\cmidrule{2-7}
 & \multirow[c]{2}{*}{\texttt{sales-incoterms}} & Val & $31.83$ & $61.00$ & $60.53_{\pm 0.09}$ & $\bm{69.07}_{\pm 1.46}$ \\
 &  & Test & $29.39$ & $56.63$ & $56.63_{\pm 0.00}$ & $\bm{62.23}_{\pm 0.53}$ \\
\cmidrule{2-7}
 & \multirow[c]{2}{*}{\texttt{sales-office}} & Val & $50.01$ & $\bm{99.91}$ & $99.90_{\pm 0.00}$ & $99.91_{\pm 0.00}$ \\
 &  & Test & $49.71$ & $\bm{99.88}$ & $59.93_{\pm 54.71}$ & $99.88_{\pm 0.00}$ \\
\cmidrule{2-7}
 & \multirow[c]{2}{*}{\texttt{sales-payterms}} & Val & $0.32$ & $0.65$ & $1.85_{\pm 0.33}$ & $\bm{39.88}_{\pm 0.19}$ \\
 &  & Test & $0.24$ & $0.47$ & $5.64_{\pm 5.14}$ & $\bm{37.47}_{\pm 0.43}$ \\
\cmidrule{2-7}
 & \multirow[c]{2}{*}{\texttt{sales-shipcond}} & Val & $16.49$ & $27.61$ & $31.92_{\pm 0.04}$ & $\bm{59.21}_{\pm 1.87}$ \\
 &  & Test & $15.64$ & $26.30$ & $4.91_{\pm 0.00}$ & $\bm{56.85}_{\pm 1.32}$ \\
\cmidrule{1-7} \cmidrule{2-7}
\multirow[c]{2}{*}{\texttt{rel-stack}} & \multirow[c]{2}{*}{\texttt{badges-class}} & Val & $11.59$ & $20.68$ & $1.93_{\pm 0.13}$ & $\bm{79.97}_{\pm 0.05}$ \\
 &  & Test & $10.49$ & $18.34$ & $2.51_{\pm 0.00}$ & $\bm{82.83}_{\pm 0.18}$ \\
\bottomrule
\end{tabular}

%% file: tables/results/autocomplete_regression_r2_appendix.tex
\begin{tabular}{llllllllll}
\toprule
 &  &  & Zero & Mean & Median & Ent. Mean & Ent. Med. & LightGBM & GNN \\
\midrule
\multirow[c]{2}{*}{\texttt{rel-amazon}} & \multirow[c]{2}{*}{\texttt{review-rating}} & Val & $-20.848$ & $\bm{-0.006}$ & $-0.364$ & $-20.848$ & $-20.848$ & $-0.364_{\pm 0.000}$ & $-0.356_{\pm 0.012}$ \\
 &  & Test & $-22.579$ & $\bm{-0.014}$ & $-0.341$ & $-13.313$ & $-13.313$ & $-0.341_{\pm 0.000}$ & $-0.331_{\pm 0.013}$ \\
\cmidrule{1-10} \cmidrule{2-10}
\multirow[c]{2}{*}{\texttt{rel-event}} & \multirow[c]{2}{*}{\texttt{users-birthyear}} & Val & $-55012.323$ & $-0.047$ & $-0.216$ & $-55012.323$ & $-55012.323$ & $0.004_{\pm 0.004}$ & $\bm{0.008}_{\pm 0.036}$ \\
 &  & Test & $-64803.758$ & $-0.121$ & $-0.395$ & $-64803.758$ & $-64803.758$ & $-0.192_{\pm 0.108}$ & $\bm{-0.030}_{\pm 0.040}$ \\
\cmidrule{1-10} \cmidrule{2-10}
 \multirow[c]{4}{*}{\texttt{rel-f1}} & \multirow[c]{2}{*}{\texttt{qualifying-position}} & Val & $-3.267$ & $-0.018$ & $-0.030$ & $-3.267$ & $-3.267$ & $\bm{0.153}_{\pm 0.003}$ & $0.015_{\pm 0.009}$ \\
 &  & Test & $-3.214$ & $-0.002$ & $-0.001$ & $-3.214$ & $-3.214$ & $-0.953_{\pm 0.039}$ & $\bm{0.015}_{\pm 0.010}$ \\
\cmidrule{2-10}
 & \multirow[c]{2}{*}{\texttt{results-position}} & Val & $-3.123$ & $-0.100$ & $-0.107$ & $-3.123$ & $-3.123$ & $0.283_{\pm 0.006}$ & $\bm{0.440}_{\pm 0.026}$ \\
 &  & Test & $-3.148$ & $-0.176$ & $-0.219$ & $-3.148$ & $-3.148$ & $-2.437_{\pm 0.053}$ & $\bm{0.394}_{\pm 0.039}$ \\
\cmidrule{1-10} \cmidrule{2-10}
\multirow[c]{2}{*}{\texttt{rel-hm}} & \multirow[c]{2}{*}{\texttt{transactions-price}} & Val & $-2.215$ & $-0.065$ & $-0.140$ & $-2.215$ & $-2.215$ & $-0.140_{\pm 0.000}$ & $\bm{0.725}_{\pm 0.003}$ \\
 &  & Test & $-2.329$ & $-0.075$ & $-0.159$ & $-2.329$ & $-2.329$ & $-0.160_{\pm 0.000}$ & $\bm{0.736}_{\pm 0.002}$ \\
\cmidrule{1-10} \cmidrule{2-10}
\multirow[c]{2}{*}{\texttt{rel-ratebeer}} & \multirow[c]{2}{*}{\texttt{user-beer-rating}} & Val & $-23.411$ & $-0.015$ & $-0.004$ & $-23.411$ & $-23.411$ & $-0.004_{\pm 0.000}$ & $\bm{0.448}_{\pm 0.006}$ \\
 &  & Test & $-34.352$ & $-0.031$ & $-0.003$ & $-34.352$ & $-34.352$ & $-0.014_{\pm 0.000}$ & $\bm{0.394}_{\pm 0.010}$ \\
\cmidrule{1-10} \cmidrule{2-10}
\multirow[c]{2}{*}{\texttt{rel-trial}} & \multirow[c]{2}{*}{\texttt{studies-enrollment}} & Val & $-0.001$ & $\bm{-0.000}$ & $-0.001$ & $-0.001$ & $-0.001$ & $-0.001_{\pm 0.000}$ & $-0.000_{\pm 0.000}$ \\
 &  & Test & $-0.000$ & $\bm{-0.000}$ & $-0.000$ & $-0.000$ & $-0.000$ & $-0.000_{\pm 0.000}$ & $-0.000_{\pm 0.000}$ \\
\cmidrule{1-10} \cmidrule{2-10}
\bottomrule
\end{tabular}

%% file: tables/results/autocomplete_regression_appendix.tex
\begin{tabular}{lllrrrrrrr}
\toprule
Dataset & Task & Split & Zero & Mean & Median & Ent. Mean & Ent. Med. & LightGBM & GNN \\
\midrule
\multirow[c]{2}{*}{\texttt{rel-amazon}} & \multirow[c]{2}{*}{\texttt{review-rating}} & Val & $4.416$ & $0.779$ & $\bm{0.584}$ & $4.416$ & $4.416$ & $0.584_{\pm 0.000}$ & $0.586_{\pm 0.003}$ \\
 &  & Test & $4.453$ & $0.762$ & $\bm{0.547}$ & $2.907$ & $2.907$ & $0.547_{\pm 0.000}$ & $0.549_{\pm 0.004}$ \\
\cmidrule{1-10} \cmidrule{2-10}
\multirow[c]{2}{*}{\texttt{rel-event}} & \multirow[c]{2}{*}{\texttt{users-birthyear}} & Val & $1987.058$ & $5.668$ & $5.885$ & $1987.058$ & $1987.058$ & $\bm{5.144}_{\pm 0.006}$ & $5.193_{\pm 0.044}$ \\
 &  & Test & $1986.094$ & $5.588$ & $6.198$ & $1986.094$ & $1986.094$ & $5.752_{\pm 0.164}$ & $\bm{5.224}_{\pm 0.046}$ \\
\cmidrule{1-10} \cmidrule{2-10}
 \multirow[c]{8}{*}{\texttt{rel-f1}} & \multirow[c]{2}{*}{\texttt{qualifying-position}} & Val & $10.943$ & $5.266$ & $5.280$ & $10.943$ & $10.943$ & $\bm{4.411}_{\pm 0.014}$ & $5.190_{\pm 0.023}$ \\
 &  & Test & $11.142$ & $5.358$ & $5.342$ & $11.142$ & $11.142$ & $7.048_{\pm 0.067}$ & $\bm{5.312}_{\pm 0.023}$ \\
\cmidrule{2-10}
 & \multirow[c]{2}{*}{\texttt{results-position}} & Val & $8.587$ & $4.251$ & $4.257$ & $8.587$ & $8.587$ & $3.167_{\pm 0.014}$ & $\bm{2.779}_{\pm 0.069}$ \\
 &  & Test & $9.504$ & $4.822$ & $4.877$ & $9.504$ & $9.504$ & $8.377_{\pm 0.080}$ & $\bm{3.139}_{\pm 0.104}$ \\
\cmidrule{1-10} \cmidrule{2-10}
\multirow[c]{2}{*}{\texttt{rel-hm}} & \multirow[c]{2}{*}{\texttt{transactions-price}} & Val & $0.034$ & $0.015$ & $0.015$ & $0.034$ & $0.034$ & $0.015_{\pm 0.000}$ & $\bm{0.004}_{\pm 0.000}$ \\
 &  & Test & $0.034$ & $0.015$ & $0.015$ & $0.034$ & $0.034$ & $0.015_{\pm 0.000}$ & $\bm{0.004}_{\pm 0.000}$ \\
\cmidrule{1-10} \cmidrule{2-10}
\multirow[c]{2}{*}{\texttt{rel-ratebeer}} & \multirow[c]{2}{*}{\texttt{beer\_ratings-total\_score}} & Val & $3.444$ & $0.530$ & $0.520$ & $3.444$ & $3.444$ & $0.520_{\pm 0.000}$ & $\bm{0.355}_{\pm 0.001}$ \\
 &  & Test & $3.470$ & $0.457$ & $0.431$ & $3.470$ & $3.470$ & $0.447_{\pm 0.000}$ & $\bm{0.323}_{\pm 0.003}$ \\
\cmidrule{1-10} \cmidrule{2-10}
\multirow[c]{2}{*}{\texttt{rel-trial}} & \multirow[c]{2}{*}{\texttt{studies-enrollment}} & Val & $3782.463$ & $5893.531$ & $3754.731$ & $3782.463$ & $3782.463$ & $3734.843_{\pm 1.368}$ & $\bm{3709.484}_{\pm 1.698}$ \\
 &  & Test & $17660.073$ & $19949.874$ & $17635.281$ & $17660.073$ & $17660.073$ & $17770.098_{\pm 72.701}$ & $\bm{17604.633}_{\pm 1.562}$ \\
\bottomrule
\end{tabular}

%% file: tables/results/entity_classification_appendix.tex
\begin{tabular}{lllrr}
\toprule
Dataset & Task & Split & LightGBM & GNN \\
\midrule
\multirow[c]{2}{*}{\texttt{rel-arxiv}} & \multirow[c]{2}{*}{\texttt{paper-citation}} & Val & $71.94_{\pm 0.12}$ & $\bm{82.45}_{\pm 0.03}$ \\
 &  & Test & $71.21_{\pm 0.13}$ & $\bm{82.50}_{\pm 0.04}$ \\
\cmidrule{1-5} \cmidrule{2-5}
\multirow[c]{2}{*}{\texttt{rel-mimic}} & \multirow[c]{2}{*}{\texttt{patient-iculengthofstay}} & Val & $53.64_{\pm 0.28}$ & $\bm{56.52}_{\pm 0.05}$ \\
 &  & Test & $51.81_{\pm 0.56}$ & $\bm{55.01}_{\pm 0.11}$ \\
\cmidrule{1-5} \cmidrule{2-5}
\multirow[c]{6}{*}{\texttt{rel-ratebeer}} & \multirow[c]{2}{*}{\texttt{beer-churn}} & Val & $81.90_{\pm 0.06}$ & $\bm{90.47}_{\pm 0.06}$ \\
 &  & Test & $76.21_{\pm 0.12}$ & $\bm{78.67}_{\pm 0.60}$ \\
\cmidrule{2-5}
 & \multirow[c]{2}{*}{\texttt{brewer-dormant}} & Val & $76.39_{\pm 0.10}$ & $\bm{82.10}_{\pm 0.13}$ \\
 &  & Test & $75.79_{\pm 0.11}$ & $\bm{80.51}_{\pm 0.17}$ \\
\cmidrule{2-5}
 & \multirow[c]{2}{*}{\texttt{user-churn}} & Val & $87.02_{\pm 0.02}$ & $\bm{96.85}_{\pm 0.26}$ \\
 &  & Test & $83.92_{\pm 18.42}$ & $\bm{94.27}_{\pm 0.22}$ \\
\bottomrule
\end{tabular}

%% file: tables/results/entity_multiclass_appendix.tex
\begin{tabular}{lllrrrr}
\toprule
Dataset & Task & Split & Random & Majority & LightGBM & GNN \\
\midrule
\multirow[c]{2}{*}{\texttt{rel-arxiv}} & \multirow[c]{2}{*}{\texttt{author-category}} & Val & $1.75$ & $8.83$ & $1.95_{\pm 0.16}$ & $\bm{52.63}_{\pm 0.08}$ \\
 &  & Test & $1.77$ & $9.09$ & $2.01_{\pm 0.21}$ & $\bm{50.74}_{\pm 1.01}$ \\
\bottomrule
\end{tabular}

%% file: tables/results/entity_regression_r2_appendix.tex
\begin{tabular}{llllllllll}
\toprule
 &  &  & Zero & Mean & Median & Ent. Mean & Ent. Med. & LightGBM & GNN \\
\midrule
\multirow[c]{4}{*}{\texttt{rel-amazon}} & \multirow[c]{2}{*}{\texttt{item-ltv}} & Val & $-0.025$ & $-0.000$ & $-0.013$ & $-0.247$ & $-0.107$ & $0.002_{\pm 0.001}$ & $\bm{0.066}_{\pm 0.003}$ \\
 &  & Test & $-0.013$ & $-0.000$ & $-0.007$ & $0.030$ & $\bm{0.099}$ & $0.001_{\pm 0.000}$ & $0.032_{\pm 0.001}$ \\
\cmidrule{2-10}
 & \multirow[c]{2}{*}{\texttt{user-ltv}} & Val & $-0.084$ & $-0.003$ & $-0.084$ & $0.053$ & $0.095$ & $-0.084_{\pm 0.000}$ & $\bm{0.195}_{\pm 0.006}$ \\
 &  & Test & $-0.092$ & $-0.000$ & $-0.092$ & $0.143$ & $0.168$ & $-0.092_{\pm 0.000}$ & $\bm{0.172}_{\pm 0.006}$ \\
\cmidrule{1-10} \cmidrule{2-10}
\multirow[c]{2}{*}{\texttt{rel-arxiv}} & \multirow[c]{2}{*}{\texttt{author-publication}} & Val & $-1.579$ & $-0.012$ & $-0.259$ & $0.254$ & $0.236$ & $-0.259_{\pm 0.000}$ & $\bm{0.437}_{\pm 0.013}$ \\
 &  & Test & $-1.572$ & $-0.000$ & $-0.210$ & $-0.010$ & $-0.064$ & $-0.210_{\pm 0.000}$ & $\bm{0.249}_{\pm 0.013}$ \\
\cmidrule{1-10} \cmidrule{2-10}
\multirow[c]{2}{*}{\texttt{rel-avito}} & \multirow[c]{2}{*}{\texttt{ad-ctr}} & Val & $-0.238$ & $-0.002$ & $-0.095$ & $-0.224$ & $-0.224$ & $-0.032_{\pm 0.005}$ & $\bm{0.030}_{\pm 0.017}$ \\
 &  & Test & $-0.226$ & $-0.004$ & $-0.098$ & $-0.148$ & $-0.148$ & $-0.039_{\pm 0.004}$ & $\bm{-0.001}_{\pm 0.020}$ \\
\cmidrule{1-10} \cmidrule{2-10}
\multirow[c]{2}{*}{\texttt{rel-event}} & \multirow[c]{2}{*}{\texttt{user-attendance}} & Val & $-0.249$ & $\bm{-0.037}$ & $-0.249$ & $-0.193$ & $-0.147$ & $-0.249_{\pm 0.001}$ & $-0.045_{\pm 0.116}$ \\
 &  & Test & $-0.168$ & $-0.019$ & $-0.168$ & $-0.065$ & $-0.043$ & $-0.168_{\pm 0.000}$ & $\bm{0.003}_{\pm 0.096}$ \\
\cmidrule{1-10} \cmidrule{2-10}
\multirow[c]{2}{*}{\texttt{rel-f1}} & \multirow[c]{2}{*}{\texttt{driver-position}} & Val & $-5.715$ & $-0.370$ & $-0.236$ & $-2.866$ & $-2.840$ & $0.150_{\pm 0.025}$ & $\bm{0.249}_{\pm 0.008}$ \\
 &  & Test & $-5.239$ & $-0.119$ & $-0.042$ & $-2.841$ & $-2.849$ & $\bm{0.068}_{\pm 0.049}$ & $0.039_{\pm 0.063}$ \\
\cmidrule{1-10} \cmidrule{2-10}
\multirow[c]{2}{*}{\texttt{rel-hm}} & \multirow[c]{2}{*}{\texttt{item-sales}} & Val & $-0.017$ & $-0.000$ & $-0.017$ & $0.065$ & $0.053$ & $-0.017_{\pm 0.000}$ & $\bm{0.187}_{\pm 0.002}$ \\
 &  & Test & $-0.017$ & $-0.000$ & $-0.017$ & $0.058$ & $0.042$ & $-0.017_{\pm 0.000}$ & $\bm{0.215}_{\pm 0.002}$ \\
\cmidrule{1-10} \cmidrule{2-10}
\multirow[c]{2}{*}{\texttt{rel-ratebeer}} & \multirow[c]{2}{*}{\texttt{user-count}} & Val & $-0.037$ & $-0.053$ & $-0.037$ & $0.551$ & $0.547$ & $\bm{0.559}_{\pm 0.014}$ & $0.526_{\pm 0.005}$ \\
 &  & Test & $-0.071$ & $-0.025$ & $-0.071$ & $0.264$ & $0.285$ & $-0.170_{\pm 0.684}$ & $\bm{0.625}_{\pm 0.003}$ \\
\cmidrule{1-10} \cmidrule{2-10}
\multirow[c]{2}{*}{\texttt{rel-stack}} & \multirow[c]{2}{*}{\texttt{post-votes}} & Val & $-0.028$ & $-0.007$ & $-0.028$ & $\bm{0.306}$ & $0.285$ & $-0.028_{\pm 0.000}$ & $0.122_{\pm 0.003}$ \\
 &  & Test & $-0.034$ & $-0.004$ & $-0.034$ & $\bm{0.294}$ & $0.272$ & $-0.034_{\pm 0.000}$ & $0.122_{\pm 0.004}$ \\
\cmidrule{1-10} \cmidrule{2-10}
\multirow[c]{4}{*}{\texttt{rel-trial}} & \multirow[c]{2}{*}{\texttt{site-success}} & Val & $-0.988$ & $\bm{-0.005}$ & $-0.988$ & $-0.749$ & $-0.809$ & $-0.319_{\pm 0.077}$ & $-0.425_{\pm 0.059}$ \\
 &  & Test & $-0.923$ & $\bm{-0.001}$ & $-0.923$ & $-0.714$ & $-0.751$ & $-0.336_{\pm 0.087}$ & $-0.483_{\pm 0.110}$ \\
\cmidrule{2-10}
 & \multirow[c]{2}{*}{\texttt{study-adverse}} & Val & $-0.021$ & $-0.002$ & $-0.020$ & $-0.021$ & $-0.021$ & $\bm{0.134}_{\pm 0.017}$ & $0.066_{\pm 0.003}$ \\
 &  & Test & $-0.054$ & $-0.005$ & $-0.050$ & $-0.054$ & $-0.054$ & $\bm{0.307}_{\pm 0.039}$ & $0.177_{\pm 0.009}$ \\
\cmidrule{1-10} \cmidrule{2-10}
\bottomrule
\end{tabular}

%% file: tables/results/entity_regression_appendix.tex
\begin{tabular}{lllrrrrrrr}
\toprule
Dataset & Task & Split & Zero & Mean & Median & Ent. Mean & Ent. Med. & LightGBM & GNN \\
\midrule
\multirow[c]{2}{*}{\texttt{rel-arxiv}} & \multirow[c]{2}{*}{\texttt{author-publication}} & Val & $1.681$ & $0.864$ & $0.681$ & $0.827$ & $0.804$ & $0.681_{\pm 0.000}$ & $\bm{0.435}_{\pm 0.008}$ \\
 &  & Test & $1.577$ & $0.769$ & $0.577$ & $0.879$ & $0.874$ & $0.577_{\pm 0.000}$ & $\bm{0.513}_{\pm 0.008}$ \\
\cmidrule{1-10} \cmidrule{2-10}
\multirow[c]{2}{*}{\texttt{rel-ratebeer}} & \multirow[c]{2}{*}{\texttt{user-count}} & Val & $11.255$ & $28.892$ & $11.255$ & $8.363$ & $7.866$ & $7.065_{\pm 0.058}$ & $\bm{5.813}_{\pm 0.031}$ \\
 &  & Test & $15.124$ & $29.050$ & $15.124$ & $13.883$ & $13.079$ & $20.350_{\pm 9.536}$ & $\bm{7.374}_{\pm 0.102}$ \\
\bottomrule
\end{tabular}

%% file: tables/results/recommendation_appendix.tex
\begin{tabular}{lllrrrrrrr}
\toprule
Dataset & Task & Split & Past Visit & Global Pop. & LightGBM & GNN (2) & GNN (4) & IDGNN (2) & IDGNN (4) \\
\midrule
\multirow[c]{6}{*}{\texttt{rel-amazon}} & \multirow[c]{2}{*}{\texttt{user-item-purchase}} & Val & $0.07$ & $0.31$ & $0.16_{\pm 0.02}$ & $1.56_{\pm 0.05}$ & $\bm{1.60}_{\pm 0.07}$ & $0.13_{\pm 0.00}$ & $0.13_{\pm 0.00}$ \\
 &  & Test & $0.06$ & $0.24$ & $0.14_{\pm 0.01}$ & $0.74_{\pm 0.06}$ & $\bm{0.91}_{\pm 0.09}$ & $0.10_{\pm 0.00}$ & $0.11_{\pm 0.00}$ \\
\cmidrule{2-10}
 & \multirow[c]{2}{*}{\texttt{user-item-rate}} & Val & $0.09$ & $0.16$ & $0.23_{\pm 0.04}$ & $1.42_{\pm 0.13}$ & $\bm{1.54}_{\pm 0.08}$ & $0.15_{\pm 0.00}$ & $0.15_{\pm 0.00}$ \\
 &  & Test & $0.07$ & $0.15$ & $0.18_{\pm 0.02}$ & $0.81_{\pm 0.06}$ & $\bm{0.98}_{\pm 0.09}$ & $0.12_{\pm 0.00}$ & $0.12_{\pm 0.00}$ \\
\cmidrule{2-10}
 & \multirow[c]{2}{*}{\texttt{user-item-review}} & Val & $0.05$ & $0.18$ & $0.16_{\pm 0.03}$ & $1.02_{\pm 0.03}$ & $\bm{1.19}_{\pm 0.06}$ & $0.11_{\pm 0.00}$ & $0.11_{\pm 0.00}$ \\
 &  & Test & $0.04$ & $0.11$ & $0.10_{\pm 0.01}$ & $0.46_{\pm 0.02}$ & $\bm{0.63}_{\pm 0.03}$ & $0.09_{\pm 0.00}$ & $0.09_{\pm 0.00}$ \\
\cmidrule{1-10} \cmidrule{2-10}
\multirow[c]{2}{*}{\texttt{rel-avito}} & \multirow[c]{2}{*}{\texttt{user-ad-visit}} & Val & $3.66$ & $0.01$ & $0.17_{\pm 0.02}$ & $0.10_{\pm 0.03}$ & $0.12_{\pm 0.01}$ & $5.40_{\pm 0.02}$ & $\bm{5.78}_{\pm 0.04}$ \\
 &  & Test & $1.95$ & $0.00$ & $0.06_{\pm 0.01}$ & $0.04_{\pm 0.01}$ & $0.05_{\pm 0.01}$ & $3.65_{\pm 0.02}$ & $\bm{3.92}_{\pm 0.03}$ \\
\cmidrule{1-10} \cmidrule{2-10}
\multirow[c]{2}{*}{\texttt{rel-hm}} & \multirow[c]{2}{*}{\texttt{user-item-purchase}} & Val & $1.07$ & $0.36$ & $0.43_{\pm 0.03}$ & $0.87_{\pm 0.06}$ & $0.68_{\pm 0.04}$ & $2.64_{\pm 0.01}$ & $\bm{2.70}_{\pm 0.01}$ \\
 &  & Test & $0.89$ & $0.30$ & $0.37_{\pm 0.03}$ & $0.80_{\pm 0.07}$ & $0.68_{\pm 0.06}$ & $2.80_{\pm 0.01}$ & $\bm{2.91}_{\pm 0.01}$ \\
\cmidrule{1-10} \cmidrule{2-10}
\multirow[c]{4}{*}{\texttt{rel-stack}}  & \multirow[c]{2}{*}{\texttt{user-post-comment}} & Val & $2.05$ & $0.03$ & $0.04_{\pm 0.02}$ & $0.40_{\pm 0.09}$ & $0.56_{\pm 0.08}$ & $15.17_{\pm 0.15}$ & $\bm{15.60}_{\pm 0.10}$ \\
 &  & Test & $1.42$ & $0.02$ & $0.03_{\pm 0.01}$ & $0.16_{\pm 0.06}$ & $0.16_{\pm 0.03}$ & $12.72_{\pm 0.22}$ & $\bm{13.75}_{\pm 0.36}$ \\
\cmidrule{2-10}
 & \multirow[c]{2}{*}{\texttt{post-post-related}} & Val & $0.00$ & $0.47$ & $0.86_{\pm 0.17}$ & $0.02_{\pm 0.03}$ & $0.00_{\pm 0.00}$ & $7.83_{\pm 0.17}$ & $\bm{8.35}_{\pm 0.22}$ \\
 &  & Test & $1.74$ & $1.46$ & $1.86_{\pm 0.56}$ & $0.03_{\pm 0.02}$ & $0.05_{\pm 0.03}$ & $10.71_{\pm 0.30}$ & $\bm{12.46}_{\pm 0.36}$ \\
\cmidrule{1-10} \cmidrule{2-10}
\multirow[c]{4}{*}{\texttt{rel-trial}} & \multirow[c]{2}{*}{\texttt{condition-sponsor-run}} & Val & $8.58$ & $2.63$ & $4.64_{\pm 0.27}$ & $3.11_{\pm 0.31}$ & $2.84_{\pm 0.37}$ & $\bm{11.33}_{\pm 0.04}$ & $\bm{11.33}_{\pm 0.07}$ \\
 &  & Test & $8.42$ & $2.52$ & $4.53_{\pm 0.21}$ & $3.07_{\pm 0.12}$ & $2.69_{\pm 0.40}$ & $\bm{11.36}_{\pm 0.04}$ & $11.31_{\pm 0.23}$ \\
\cmidrule{2-10}
 & \multirow[c]{2}{*}{\texttt{site-sponsor-run}} & Val & $15.90$ & $4.91$ & $10.69_{\pm 0.69}$ & $14.17_{\pm 0.37}$ & $14.09_{\pm 0.46}$ & $\bm{17.43}_{\pm 0.07}$ & $17.41_{\pm 0.05}$ \\
 &  & Test & $17.31$ & $3.75$ & $8.21_{\pm 0.54}$ & $10.37_{\pm 0.92}$ & $11.05_{\pm 0.94}$ & $19.00_{\pm 0.12}$ & $\bm{19.04}_{\pm 0.05}$ \\
\cmidrule{1-10} \cmidrule{2-10}

\multirow[c]{2}{*}{\texttt{rel-arxiv}} & \multirow[c]{2}{*}{\texttt{paper-paper-cocitation}} & Val & $19.01$ & $1.25$ & $12.49_{\pm 0.60}$ & $12.19_{\pm 0.23}$ & $12.96_{\pm 00.33}$ & $25.22_{\pm 0.09}$ & $\bm{35.76}_{\pm 0.09}$ \\
 &  & Test & $16.51$ & $1.13$ & $11.01_{\pm 0.43}$ & $8.83_{\pm 0.38}$ & $10.46_{\pm 0.56}$ & $22.95_{\pm 0.07}$ & $\bm{35.39}_{\pm 0.17}$ \\
\cmidrule{1-10} \cmidrule{2-10}
\multirow[c]{2}{*}{\texttt{rel-f1}} & \multirow[c]{2}{*}{\texttt{driver-circuit-compete}} & Val & $53.41$ & $55.19$ & $66.06_{\pm 2.68}$ & $3.60_{\pm 2.11}$ & $10.57_{\pm 8.35}$ & $70.70_{\pm 0.00}$ & $\bm{74.40}_{\pm 1.03}$ \\
 &  & Test & $20.76$ & $50.12$ & $57.77_{\pm 2.95}$ & $9.67_{\pm 10.56}$ & $16.57_{\pm 11.17}$ & $62.32_{\pm 0.00}$ & $\bm{76.18}_{\pm 6.59}$ \\
\cmidrule{1-10} \cmidrule{2-10}
\multirow[c]{6}{*}{\texttt{rel-ratebeer}} & \multirow[c]{2}{*}{\texttt{user-beer-favorite}} & Val & $0.00$ & $2.33$ & $1.24_{\pm 0.15}$ & $2.09_{\pm 0.15}$ & $-$ & $3.09_{\pm 0.14}$ & $\bm{3.33}_{\pm 0.09}$ \\
 &  & Test & $0.00$ & $1.10$ & $0.67_{\pm 0.13}$ & $0.56_{\pm 0.22}$ & $-$ & $1.21_{\pm 0.93}$ & $\bm{1.89}_{\pm 0.09}$ \\
\cmidrule{2-10}
 & \multirow[c]{2}{*}{\texttt{user-beer-liked}} & Val & $0.00$ & $0.77$ & $0.43_{\pm 0.02}$ & $0.77_{\pm 0.06}$ & $-$ & $0.21_{\pm 0.01}$ & $\bm{1.48}_{\pm 0.06}$ \\
 &  & Test & $0.00$ & $0.61$ & $0.29_{\pm 0.04}$ & $0.54_{\pm 0.22}$ & $-$ & $0.32_{\pm 0.03}$ & $\bm{1.46}_{\pm 0.19}$ \\
\cmidrule{2-10}
 & \multirow[c]{2}{*}{\texttt{user-place-liked}} & Val & $0.00$ & $0.24$ & $0.24_{\pm 0.07}$ & $1.06_{\pm 0.17}$ & $-$ & $0.88_{\pm 0.09}$ & $\bm{2.20}_{\pm 0.24}$ \\
 &  & Test & $0.00$ & $0.11$ & $0.08_{\pm 0.03}$ & $1.15_{\pm 0.34}$ & $-$ & $0.60_{\pm 0.50}$ & $\bm{1.85}_{\pm 0.30}$ \\
\cmidrule{1-10} \cmidrule{2-10}

\bottomrule
\end{tabular}

%% file: main.bbl
\begin{thebibliography}{34}
\providecommand{\natexlab}[1]{#1}
\providecommand{\url}[1]{\texttt{#1}}
\expandafter\ifx\csname urlstyle\endcsname\relax
  \providecommand{\doi}[1]{doi: #1}\else
  \providecommand{\doi}{doi: \begingroup \urlstyle{rm}\Url}\fi

\bibitem[Chen et~al.(2025)Chen, Kanatsoulis, and Leskovec]{chen2025relgnn}
Tianlang Chen, Charilaos Kanatsoulis, and Jure Leskovec.
\newblock {RelGNN}: Composite message passing for relational deep learning.
\newblock In \emph{International Conference on Machine Learning (ICML)}, volume 267, pp.\  8296--8312. PMLR, 2025.

\bibitem[Dwivedi et~al.(2025)Dwivedi, Kanatsoulis, Huang, and Leskovec]{dwivedi2025relational2}
Vijay~Prakash Dwivedi, Charilaos Kanatsoulis, Shenyang Huang, and Jure Leskovec.
\newblock Relational deep learning: Challenges, foundations and next-generation architectures.
\newblock In \emph{ACM SIGKDD Conference on Knowledge Discovery and Data Mining (KDD)}, volume~2, pp.\  5999--6009. ACM, 2025.

\bibitem[Dwivedi et~al.(2026)Dwivedi, Jaladi, Shen, L{\'o}pez, Kanatsoulis, Puri, Fey, and Leskovec]{dwivedi2025relational}
Vijay~Prakash Dwivedi, Sri Jaladi, Yangyi Shen, Federico L{\'o}pez, Charilaos~I Kanatsoulis, Rishi Puri, Matthias Fey, and Jure Leskovec.
\newblock Relational graph transformer.
\newblock In \emph{International Conference on Learning Representations (ICLR)}, 2026.

\bibitem[Fey \& Lenssen(2019)Fey and Lenssen]{fey2019fast}
Matthias Fey and Jan~Eric Lenssen.
\newblock Fast graph representation learning with pytorch geometric.
\newblock In \emph{ICLR 2019 Workshop on Representation Learning on Graphs and Manifolds}, 2019.
\newblock URL \url{https://rlgm.github.io/papers/2.pdf}.

\bibitem[Fey et~al.(2024)Fey, Hu, Huang, Lenssen, Ranjan, Robinson, Ying, You, and Leskovec]{rdl}
Matthias Fey, Weihua Hu, Kexin Huang, Jan~Eric Lenssen, Rishabh Ranjan, Joshua Robinson, Rex Ying, Jiaxuan You, and Jure Leskovec.
\newblock {Position}: Relational deep learning-graph representation learning on relational databases.
\newblock In \emph{International Conference on Machine Learning (ICML)}, volume 235, pp.\  13592--13607. PMLR, 2024.

\bibitem[Fey et~al.(2025)Fey, Kocijan, Lopez, Lenssen, and Leskovec]{kumorfm}
Matthias Fey, Vid Kocijan, Federico Lopez, Jan~Eric Lenssen, and Jure Leskovec.
\newblock {KumoRFM}: A foundation model for in-context learning on relational data, 2025.

\bibitem[Gorishniy et~al.(2021)Gorishniy, Rubachev, Khrulkov, and Babenko]{gorishniy2021revisiting}
Yury Gorishniy, Ivan Rubachev, Valentin Khrulkov, and Artem Babenko.
\newblock Revisiting deep learning models for tabular data.
\newblock In \emph{Advances in Neural Information Processing Systems (NeurIPS)}, volume~34, pp.\  18932--18943, 2021.

\bibitem[Hamilton et~al.(2017)Hamilton, Ying, and Leskovec]{hamilton2017inductive}
Will Hamilton, Zhitao Ying, and Jure Leskovec.
\newblock Inductive representation learning on large graphs.
\newblock In \emph{Advances in Neural Information Processing Systems (NeurIPS)}, volume~30, 2017.

\bibitem[Hanley \& McNeil(1983)Hanley and McNeil]{hanley1983method}
James~A Hanley and Barbara~J McNeil.
\newblock A method of comparing the areas under receiver operating characteristic curves derived from the same cases.
\newblock \emph{Radiology}, 148\penalty0 (3):\penalty0 839--843, 1983.

\bibitem[Hollmann et~al.(2023)Hollmann, M{\"u}ller, Eggensperger, and Hutter]{tabpfn}
Noah Hollmann, Samuel M{\"u}ller, Katharina Eggensperger, and Frank Hutter.
\newblock Tab{PFN}: A transformer that solves small tabular classification problems in a second.
\newblock In \emph{International Conference on Learning Representations (ICLR)}, 2023.

\bibitem[Hollmann et~al.(2025)Hollmann, M{\"u}ller, Purucker, Krishnakumar, K{\"o}rfer, Hoo, Schirrmeister, and Hutter]{tabpfnv2}
Noah Hollmann, Samuel M{\"u}ller, Lennart Purucker, Arjun Krishnakumar, Max K{\"o}rfer, Shi~Bin Hoo, Robin~Tibor Schirrmeister, and Frank Hutter.
\newblock Accurate predictions on small data with a tabular foundation model.
\newblock \emph{Nature}, 637\penalty0 (8045):\penalty0 319--326, 2025.

\bibitem[Hu et~al.(2024)Hu, Yuan, Zhang, Nitta, Cao, Kocijan, Leskovec, and Fey]{hu2024pytorch}
Weihua Hu, Yiwen Yuan, Zecheng Zhang, Akihiro Nitta, Kaidi Cao, Vid Kocijan, Jure Leskovec, and Matthias Fey.
\newblock {PyTorch Frame}: A modular framework for multi-modal tabular learning, 2024.
\newblock URL \url{https://arxiv.org/abs/2404.00776}.

\bibitem[Hudovernik et~al.(2025)Hudovernik, Xu, Shi, Šubelj, Ermon, Štrumbelj, and Leskovec]{hudovernik2025reldiff}
Valter Hudovernik, Minkai Xu, Juntong Shi, Lovro Šubelj, Stefano Ermon, Erik Štrumbelj, and Jure Leskovec.
\newblock {RelDiff}: Relational data generative modeling with graph-based diffusion models, 2025.
\newblock URL \url{https://arxiv.org/abs/2506.00710}.

\bibitem[Johnson et~al.(2024)Johnson, Bulgarelli, Pollard, Gow, Moody, Horng, Celi, and Mark]{johnson2024mimic}
Alistair Johnson, Lucas Bulgarelli, Tom Pollard, Brian Gow, Benjamin Moody, Steven Horng, Leo~Anthony Celi, and Roger Mark.
\newblock {MIMIC-IV}.
\newblock \emph{{PhysioNet}}, October 2024.
\newblock \doi{10.13026/kpb9-mt58}.
\newblock URL \url{https://doi.org/10.13026/kpb9-mt58}.
\newblock Version 3.1.

\bibitem[Kanatsoulis et~al.(2025)Kanatsoulis, Choi, Jegelka, Leskovec, and Ribeiro]{kanatsoulis2025learningefficientpositionalencodings}
Charilaos Kanatsoulis, Evelyn Choi, Stefanie Jegelka, Jure Leskovec, and Alejandro Ribeiro.
\newblock Learning efficient positional encodings with graph neural networks.
\newblock In \emph{International Conference on Learning Representations (ICLR)}, 2025.

\bibitem[Kapoor \& Narayanan(2023)Kapoor and Narayanan]{kapoor2023leakage}
Sayash Kapoor and Arvind Narayanan.
\newblock Leakage and the reproducibility crisis in machine-learning-based science.
\newblock \emph{Patterns}, 4\penalty0 (9), 2023.

\bibitem[Ke et~al.(2017)Ke, Meng, Finley, Wang, Chen, Ma, Ye, and Liu]{ke2017lightgbm}
Guolin Ke, Qi~Meng, Thomas Finley, Taifeng Wang, Wei Chen, Weidong Ma, Qiwei Ye, and Tie-Yan Liu.
\newblock {LightGBM}: A highly efficient gradient boosting decision tree.
\newblock In \emph{Advances in Neural Information Processing Systems (NeurIPS)}, volume~30, 2017.

\bibitem[Ketata et~al.(2025)Ketata, L{\"u}dke, Schwinn, and G{\"u}nnemann]{ketata2025joint}
Mohamed~Amine Ketata, David L{\"u}dke, Leo Schwinn, and Stephan G{\"u}nnemann.
\newblock Joint relational database generation via graph-conditional diffusion models, 2025.
\newblock URL \url{https://arxiv.org/abs/2505.16527}.

\bibitem[Kim et~al.(2024)Kim, Grinsztajn, and Varoquaux]{carte}
Myung~Jun Kim, Leo Grinsztajn, and Gael Varoquaux.
\newblock {CARTE}: Pretraining and transfer for tabular learning.
\newblock In \emph{International Conference on Machine Learning (ICML)}, volume 235, pp.\  23843--23866. PMLR, 2024.

\bibitem[Klein et~al.(2024)Klein, Biehl, Costa, Sres, Kolk, and Hoffart]{sap-salt}
Tassilo Klein, Clemens Biehl, Margarida Costa, Andre Sres, Jonas Kolk, and Johannes Hoffart.
\newblock {SALT}: Sales autocompletion linked business tables dataset.
\newblock In \emph{NeurIPS 2024 Third Table Representation Learning Workshop}, 2024.
\newblock URL \url{https://openreview.net/forum?id=UZbELpkWIr}.

\bibitem[Kothapalli et~al.(2026)Kothapalli, Ranjan, Hudovernik, Dwivedi, Hoffart, Guestrin, and Leskovec]{kothapalli2026plurel}
Vignesh Kothapalli, Rishabh Ranjan, Valter Hudovernik, Vijay~Prakash Dwivedi, Johannes Hoffart, Carlos Guestrin, and Jure Leskovec.
\newblock {PluRel}: Synthetic data unlocks scaling laws for relational foundation models, 2026.
\newblock URL \url{https://arxiv.org/abs/2602.04029}.

\bibitem[Motl \& Schulte(2025)Motl and Schulte]{motl2025ctupraguerelationallearning}
Jan Motl and Oliver Schulte.
\newblock The {CTU} prague relational learning repository, 2025.
\newblock URL \url{https://arxiv.org/abs/1511.03086}.

\bibitem[Peleška \& Šír(2025{\natexlab{a}})Peleška and Šír]{dbformer}
Jakub Peleška and Gustav Šír.
\newblock Tabular transformers meet relational databases.
\newblock \emph{ACM Transactions on Intelligent Systems and Technology}, 16\penalty0 (5), 2025{\natexlab{a}}.

\bibitem[Peleška \& Šír(2025{\natexlab{b}})Peleška and Šír]{peleska2025redelex}
Jakub Peleška and Gustav Šír.
\newblock {REDELEX}: A framework for relational deep learning exploration.
\newblock In \emph{Joint European Conference on Machine Learning and Knowledge Discovery in Databases (ECMLPKDD)}, pp.\  438--456, 2025{\natexlab{b}}.

\bibitem[Qu et~al.(2025)Qu, Holzm{\"u}ller, Varoquaux, and Morvan]{tabicl}
Jingang Qu, David Holzm{\"u}ller, Ga{\"e}l Varoquaux, and Marine~Le Morvan.
\newblock Tab{ICL}: A tabular foundation model for in-context learning on large data.
\newblock In \emph{International Conference on Machine Learning (ICML)}, volume 267, pp.\  50817--50847, 2025.

\bibitem[Ranjan et~al.(2026)Ranjan, Hudovernik, Znidar, Kanatsoulis, Upendra, Mohammadi, Meyer, Palczewski, Guestrin, and Leskovec]{ranjan2025relational}
Rishabh Ranjan, Valter Hudovernik, Mark Znidar, Charilaos Kanatsoulis, Roshan Upendra, Mahmoud Mohammadi, Joe Meyer, Tom Palczewski, Carlos Guestrin, and Jure Leskovec.
\newblock Relational transformer: Toward zero-shot foundation models for relational data.
\newblock In \emph{International Conference on Learning Representations (ICLR)}, 2026.

\bibitem[Rendle et~al.(2012)Rendle, Freudenthaler, Gantner, and Schmidt-Thieme]{rendle2012bpr}
Steffen Rendle, Christoph Freudenthaler, Zeno Gantner, and Lars Schmidt-Thieme.
\newblock {BPR}: Bayesian personalized ranking from implicit feedback, 2012.
\newblock URL \url{https://arxiv.org/abs/1205.2618}.

\bibitem[Robinson et~al.(2024)Robinson, Ranjan, Hu, Huang, Han, Dobles, Fey, Lenssen, Yuan, Zhang, He, and Leskovec]{relbench}
Joshua Robinson, Rishabh Ranjan, Weihua Hu, Kexin Huang, Jiaqi Han, Alejandro Dobles, Matthias Fey, Jan~Eric Lenssen, Yiwen Yuan, Zecheng Zhang, Xinwei He, and Jure Leskovec.
\newblock Relbench: A benchmark for deep learning on relational databases.
\newblock In \emph{Advances in Neural Information Processing Systems (NeurIPS)}, volume~37, pp.\  21330--21341, 2024.

\bibitem[Rossi et~al.(2020)Rossi, Chamberlain, Frasca, Eynard, Monti, and Bronstein]{rossi2020temporal}
Emanuele Rossi, Ben Chamberlain, Fabrizio Frasca, Davide Eynard, Federico Monti, and Michael Bronstein.
\newblock Temporal graph networks for deep learning on dynamic graphs.
\newblock In \emph{ICML 2024 Workshop on Graph Representation Learning and Beyond}, 2020.
\newblock URL \url{https://grlplus.github.io/papers/58.pdf}.

\bibitem[Spinaci et~al.(2024)Spinaci, Polewczyk, Hoffart, Kohler, Thelin, and Klein]{portal}
Marco Spinaci, Marek Polewczyk, Johannes Hoffart, Markus~C. Kohler, Sam Thelin, and Tassilo Klein.
\newblock {PORTAL}: Scalable tabular foundation models via content-specific tokenization.
\newblock In \emph{NeurIPS 2024 Third Table Representation Learning Workshop}, 2024.
\newblock URL \url{https://openreview.net/forum?id=TSZQvknbLO}.

\bibitem[Tang et~al.(2024)Tang, He, Li, and Guo]{arxiv_physics_dataset}
Haiming Tang, Sirui He, Mengjie Li, and Zhimao Guo.
\newblock {arXiv-physics}: A large-scale physics citation and authorship dataset, 2024.
\newblock URL \url{https://github.com/PKUTHM/arxiv-physics}.

\bibitem[Wang et~al.(2024)Wang, Gan, Wipf, Cai, Li, Tang, Zhang, Zhang, Mao, Song, Wang, Li, Zhang, Yang, Qin, Lei, Zhang, Zhang, Faloutsos, and Zhang]{wang20244dbinfer}
Minjie Wang, Quan Gan, David Wipf, Zhenkun Cai, Ning Li, Jianheng Tang, Yanlin Zhang, Zizhao Zhang, Zunyao Mao, Yakun Song, Yanbo Wang, Jiahang Li, Han Zhang, Guang Yang, Xiao Qin, Chuan Lei, Muhan Zhang, Weinan Zhang, Christos Faloutsos, and Zheng Zhang.
\newblock {4DBInfer}: A {4D} benchmarking toolbox for graph-centric predictive modeling on relational {DBs}, 2024.
\newblock URL \url{https://arxiv.org/abs/2404.18209}.

\bibitem[Wang et~al.(2025)Wang, Wang, Gan, Wang, Yang, Wipf, and Zhang]{griffin}
Yanbo Wang, Xiyuan Wang, Quan Gan, Minjie Wang, Qibin Yang, David Wipf, and Muhan Zhang.
\newblock Griffin: Towards a graph-centric relational database foundation model.
\newblock In \emph{International Conference on Machine Learning (ICML)}, volume 267, pp.\  64604--64627. PMLR, 2025.

\bibitem[You et~al.(2021)You, Gomes-Selman, Ying, and Leskovec]{you2021identity}
Jiaxuan You, Jonathan~M Gomes-Selman, Rex Ying, and Jure Leskovec.
\newblock Identity-aware graph neural networks.
\newblock In \emph{Association for the Advancement of Artificial Intelligence (AAAI)}, volume~35, pp.\  10737--10745, 2021.

\end{thebibliography}
